\documentclass[10pt,journal,compsoc]{IEEEtran}
\usepackage{cite}
\usepackage{graphicx}
\usepackage{balance}  % for  \balance command ON LAST PAGE  (only there!)
\usepackage{color}
\usepackage{amsfonts,amssymb,amsmath,bm}
\usepackage{makecell}
\usepackage{subfigure}
\usepackage{multirow}
\usepackage[ruled,vlined]{algorithm2e}

\begin{document}

\title{FastSGD: A Fast Compressed SGD \\ Framework for Distributed Machine Learning}
%\title{FastSGD: Compressed Stochastic Gradient Descent for Fast Distributed Machine Learning}
%\title{FastSGD: Stochastic Gradient Descent Compression for Fast Distributed Machine Learning}

\author{Keyu Yang,~\IEEEmembership{}
        Lu Chen,~\IEEEmembership{}
        Zhihao Zeng,~\IEEEmembership{}
        Yunjun Gao,~\IEEEmembership{Member,~IEEE}

\IEEEcompsocitemizethanks{\IEEEcompsocthanksitem K. Yang, L. Chen, and Y. Gao are with the College of Computer Science, Zhejiang University, Hangzhou 310027, China.  E-mail: \{kyyang, luchen, gaoyj\}@zju.edu.cn.	
%*(Corresponding Author)
\IEEEcompsocthanksitem Z. Zeng is with the School of Software Technology, Zhejiang University, Ningbo 315048, China. E-mail: sezengzhihao@zju.edu.cn.
}

\thanks{Manuscript received xxxx, xxxx; revised xxxx, xxxx.}
}

\IEEEtitleabstractindextext{%
\begin{abstract}
With the rapid increase of big data, distributed Machine Learning (ML) has been widely applied in training large-scale models. Stochastic Gradient Descent (SGD) is arguably the workhorse algorithm of ML. Distributed ML models trained by SGD involve large amounts of gradient communication, which limits the scalability of distributed ML. Thus, it is important to compress the gradients for reducing communication. In this paper, we propose FastSGD, a Fast compressed SGD framework for distributed ML. To achieve a high compression ratio at a low cost, FastSGD represents the gradients as key-value pairs, and compresses both the gradient keys and values in linear time complexity. For the gradient value compression, FastSGD first uses a reciprocal mapper to transform original values into reciprocal values, and then, it utilizes a logarithm quantization to further reduce reciprocal values to small integers. Finally, FastSGD filters reduced gradient integers by a given threshold. For the gradient key compression, FastSGD provides an adaptive fine-grained delta encoding method to store gradient keys with fewer bits. Extensive experiments on practical ML models and datasets demonstrate that FastSGD achieves the compression ratio up to 4 orders of magnitude, and accelerates the convergence time up to 8$\times$, compared with state-of-the-art methods.
\end{abstract}

}

\maketitle

\IEEEdisplaynontitleabstractindextext

\IEEEpeerreviewmaketitle

\section{Introduction}
\label{sec:intro}

\IEEEPARstart{M}{achine} learning (ML) technology powers many aspects of modern society, such as computer vision~\cite{DBLP:conf/cvpr/DengDSLL009, DBLP:conf/cvpr/HeZRS16}, natural language processing~\cite{DBLP:journals/corr/abs-2004-03705, DBLP:journals/corr/abs-2009-06732}, speech recognition~\cite{DBLP:conf/icml/AmodeiABCCCCCCD16}, to name just a few. The performance of ML models (e.g., accuracy in classification tasks) generally improves if more training data is used~\cite{DBLP:conf/iccv/SunSSG17}. However, with the rapid increase of data, a single machine cannot support the model training process efficiently. Motivated by this, distributed training has received extensive attention from academia and industry~\cite{DBLP:journals/corr/abs-2003-06307, DBLP:conf/osdi/LiAPSAJLSS14}.

Almost all the ML models are trained with a first-order gradient descent method, namely, Stochastic Gradient Descent (SGD)~\cite{robbins1951stochastic}. To deploy the SGD-based ML algorithms in a distributed environment, the training dataset is partitioned into several distributed \textit{workers}. Each worker independently proposes local intermediate results, i.e., the gradients. Then, local gradients are communicated through the network to be aggregated, and the aggregated values are sent back to the workers for the update of the global model state. The process is repeated over many epochs  (i.e., full iterations over the entire training dataset) until the convergence.

Under the aforementioned distributed ML setting, the network communication involves large amounts of gradients, which could become the bottleneck cost to limit the scalability of the distributed system. This leads to inefficient utilization of the computational resources, longer training time, and/or higher financial cost for the cloud infrastructure. Therefore, the gradient communication should be well optimized to fully harness the computing powers of distributed workers.

Towards this, many studies have been devoted to the compression methods for the exchanged gradients before transmitting across the network with little impact on the model convergence. Existing efforts can be mainly divided into three categories: \textit{Quantization}~\cite{DBLP:conf/icml/BernsteinWAA18, DBLP:conf/nips/AlistarhG0TV17}, \textit{Sparsification}~\cite{DBLP:conf/nips/AlistarhH0KKR18, DBLP:conf/aaai/DuttaBA0SCK20}, and \textit{Hybrid} methods~\cite{DBLP:conf/sigmod/JiangFY018, DBLP:conf/mlsys/LimAK19}. Quantization methods use lower bits to represent each element in gradients (e.g., casting 32 bits to 8 bits). However, even using only one bit for each gradient, the maximal compression ratio of quantization is 32$\times$ compared to the 32-bit counterpart. In contrast, sparsification methods directly reduce the number of gradient elements to transmit (e.g., only transmitting top-$k$ largest elements). Thus, sparsification reduces the communication cost by transmitting only a small portion of gradients (e.g., 0.1\% \cite{DBLP:conf/aaai/DuttaBA0SCK20}). In addition, hybrid methods further compress the gradients by combining sparsification with quantization.

%i) \textit{Quantization} methods~\cite{DBLP:journals/corr/Dettmers15, DBLP:conf/interspeech/SeideFDLY14, DBLP:conf/icml/BernsteinWAA18, DBLP:conf/nips/AlistarhG0TV17}, which use a small number of bits to represent each element in gradients (e.g., cast 32 bits to 8 bits); ii) \textit{Sparsification} methods~\cite{DBLP:conf/nips/StichCJ18, DBLP:conf/nips/AlistarhH0KKR18, DBLP:conf/ijcai/ShiZWTC19, DBLP:conf/aaai/DuttaBA0SCK20}, which reduce the number of gradient elements to transmit (e.g., only transmit top-$k$ largest elements); iii) \textit{Hybrid} methods~\cite{DBLP:conf/nips/0001DKD19, DBLP:conf/interspeech/Strom15, DBLP:conf/sigmod/JiangFY018, DBLP:conf/mlsys/LimAK19}, which combine quantization and sparsification. In quantization methods, even using only one bit for each gradient, the maximum compression ratio is 32x compared to the 32-bit counterpart. In contrast, sparsification methods can just transmit a small portion of gradients (e.g., 0.1\%  \cite{DBLP:conf/aaai/DuttaBA0SCK20}). {\color{red}Hybrid methods further compress the gradient by combining sparsification with quantization.}
Typically, the transferred gradients are sparse originally and/or after sparsification. Nevertheless, almost all the existing gradient compression methods assume that the gradient vector to be compressed is dense. If all the dimensions of the gradient vector are stored, it would waste a lot of both storing space and processing time for zero gradient values. To this end, we can represent the nonzero elements in a gradient vector as key-value pairs. The key-value representation format of sparse gradient leaves room for further compression.

There is only one line of method SketchML~\cite{DBLP:conf/sigmod/JiangFY018,  DBLP:journals/vldb/JiangFYSC20} that provides the framework to compress both the gradient keys and values. However, SketchML is limited to the communication-intensive workloads, as it has a high computational cost for gradient compression/decompression that could be larger than the savings by the reduced communication.

Motivated by those limitations of the previous gradient compression methods, we dedicate this paper to the development of FastSGD, a \underline{Fast} compressed \underline{SGD} framework for distributed machine learning, with lightweight compression/decompression cost. In view of this, the challenges are two-fold as follows.

The first challenge is \textit{how to efficiently compress the gradient value?} The quantization and sparsification strategies inspire us that \textit{lossy} compression is suitable to reduce the volume of gradient values while guaranteeing the model convergence at the same time. FastSGD provides an effective hybrid method that combines the quantization and sparsification methods to compress the gradient values. The compression method of gradient values in FastSGD mainly contains three phases.
%adopts an effective three-phase lossy compression method for compressing the gradient values, which is a hybrid method that combines the quantization and sparsification methods.
First of all, FastSGD uses a \textit{reciprocal mapper} to transform original gradient values into reciprocal values. Then, FastSGD utilizes a \textit{logarithm quantization} method to reduce the space cost for the representation of mapped values. Finally, FastSGD adopts a \textit{threshold filter} to select the elements whose absolute values are large to transmit in the network.

The second challenge is \textit{how to efficiently compress the gradient key?} Different from the gradient values that can tolerate a precision loss during the compression/decompression, the gradient keys are vulnerable to inaccuracy. Assume that we compress a key but fail to decompress it accurately due to the precision loss, we would update a wrong gradient descent direction (i.e., a wrong dimension of gradient vector) and cannot guarantee the correct convergence of the ML model. Thus, a lossless method that transforms the keys to \textit{delta keys} has been proposed~\cite{DBLP:conf/sigmod/JiangFY018}. Nonetheless, it uses byte instead of bit as the smallest grain to encode the delta key, incurring the redundant space cost. In this paper, the proposed FastSGD goes a step further by using an adaptive fine-grained delta encoding method to store the gradient keys with fewer bits.

In brief, the key contributions of this paper are summarized as follows:

\begin{itemize}
\item{} We propose FastSGD, a fast compressed SGD framework for distributed machine learning to compress the gradients consisting of key-value pairs with lightweight compression/decompression cost.
%\item{} FastSGD compresses gradient keys and values separately, which provides a effective three-phase lossy compression for the gradient values and uses an adaptive fine-grained delta encoding method to store gradient keys with fewer bits.
\item{} FastSGD provides a hybrid lossy compression method, including the reciprocal mapper, the logarithm quantization, and the threshold filter, to compress the gradient values.
\item{} As for the compression of gradient key, FastSGD utilizes a lossless binary encoding method based on the adaptive fine-grained strategy to store the delta key with fewer bits.
\item{} We conduct extensive experiments on a range of practical ML workloads to demonstrate that FastSGD achieves up to 4 orders of magnitude compression ratio, and reduces the convergence time up to 8$\times$, compared with the state-of-the-art algorithms.
\end{itemize}

The rest of this paper is organized as follows. We introduce the distributed ML model training with SGD in Section~\ref{sec:pre}. Section~\ref{sec:method} presents the details of our proposed FastSGD. Considerable experimental results are reported in Section~\ref{sec:exp}. We review the related work in Section~\ref{sec:related}. Finally, Section~\ref{sec:conclusion} concludes the paper.

\section{Preliminaries}
\label{sec:pre}
In this section, we present some preliminary materials related to compressed SGD for distributed ML. Table~\ref{tab:symbol} summarizes the symbols used frequently.

\begin{table}
\caption{Symbols and Description}
\label{tab:symbol}
\vspace{-1mm}
\begin{tabular}{|p{1.5cm}|p{6.5cm}|}
\hline
\textbf{Notation} & \textbf{Description} \\ \hline
$x$ & the training instance \\ \hline
$y$ & the corresponding label of $x$ \\ \hline
$\theta$ & the parameters of ML model \\ \hline
$g$ & the gradient vector \\  \hline
$D_m$ & the number of model parameters \\ \hline
$D$ & the number of nonzero elements in $g$ \\ \hline
$d$ & the number of gradient elements after threshold filter \\ \hline
$(k_j, v_j)$ & the $j$-th nonzero gradient key and value in $g$ \\ \hline
$R(v_j)$ & the reciprocal gradient value for $v_j$  \\ \hline
$L(v_j)$ & the logarithm quantified gradient value for  $v_j$  \\ \hline
$\Delta k_j$ & the delta gradient key for $k_j$ \\ \hline
\end{tabular}
\end{table}

\subsection{Distributed Machine Learning}

\begin{figure}
\centering
\includegraphics[width=0.495\textwidth]{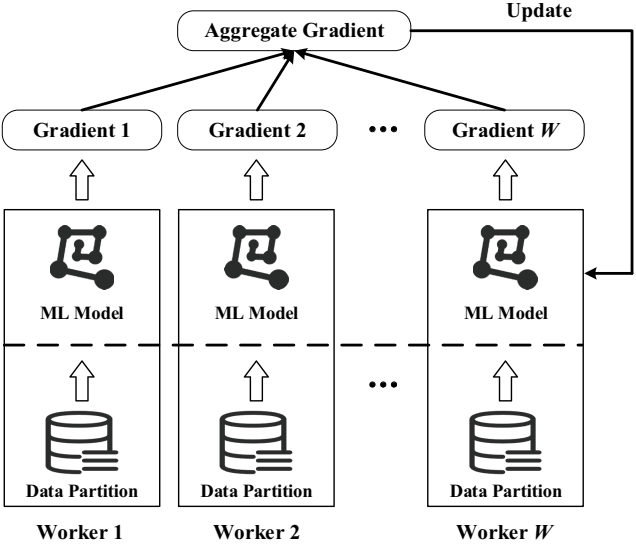}
\vspace{-2mm}
\caption{The workflow of data parallelism distributed ML}
\label{fig:disML}
\end{figure}

In this paper, we focus on \textit{data parallelism}{\footnote{\textit{Model parallelism}, which splits model parameters to multiple distributed workers~\cite{DBLP:conf/nips/DeanCMCDLMRSTYN12}, is orthogonal to data parallelism, and outside the scope of this paper.}} distributed machine learning (ML)~\cite{DBLP:journals/corr/abs-2003-06307, DBLP:conf/osdi/LiAPSAJLSS14}. The workflow of data parallelism distributed ML is illustrated in Fig.~\ref{fig:disML}. Each worker possesses a local copy of the whole model parameters and a partition of input dataset. In a single round of iteration, each worker independently processes its data to compute local updates (i.e., the local gradients), and communicates with all other workers to synchronize with the aggregated global model state.

Formally, given a set of input data $\left\{x_{i}, y_{i}\right\}_{i=1}^{N}$, i.e., the training instances $x$ and their labels $y$, and a \textit{loss} function $f(x,y,\theta)$, the distributed ML training process tries to solve the optimization problem that minimizes the loss function $f$, i.e., finding the model parameter $\theta \in \mathbb{R}^{D_m}$ that best predicts $y_{i}$ for each $x_{i}$. Here, $N$ denotes the number of training instances, and $D_m$ corresponds to the number of model parameters.

% The workflow of distributed ML is illustrated in Fig.~\ref{fig:disML}. Each worker possesses a local copy of the model parameters, processes a partition of dataset to compute local gradient updates independently, and communicates with other workers to synchronize with the aggregated global model state regularly.

%We focus on \textit{data-parallelism} distributed machine learning \cite{DBLP:journals/corr/abs-2003-06307, DBLP:conf/osdi/LiAPSAJLSS14}, in which each worker possesses a local copy of the model parameters $\theta \in \mathbb{R}^{d}$, where $d$ corresponds to the number of parameters. In a single round of iteration, each worker processes a partition of the whole training instances and their labels $\left\{x_{i}, y_{i}\right\}_{i=1}^{N}$ ($N$ is the number of training instances) to compute local updates and communicates with all other workers to synchronize with the aggregated global model state. The purpose is to find a model $\theta \in \mathbb{R}^{d}$ that minimizes a loss function

\subsection{Stochastic Gradient Descent}
Stochastic Gradient Descent (SGD) is one of the most commonly-used first-order iterative gradient descent methods~\cite{robbins1951stochastic}. At iterative round $t+1$, SGD updates the model parameters $\theta_{t+1}$ as follows:
\begin{equation}
\theta_{t+1}=\theta_{t}-\eta_{t} g_{t},
\end{equation}
where $\eta_{t}>0$ is the learning rate; $g_{t}$ is the stochastic gradient at the iterative round $t$, i.e., a unbiased estimator of the gradient of $f$; and $\theta_{t}$ denotes the model parameter values at the iterative round $t$.

Typically, the gradient $g = \nabla_\theta f(x,y,\theta) \in \mathbb{R}^{D_m}$ is a sparse vector due to the origin data distribution and/or the sparsification compression technique. To save the size of storage, we adopt the key-value pair $(k_j, v_j)$ to represent the $j$-th nonzero gradient key and value in a sparse gradient vector. Here, we denote the number of nonzero elements in the gradient as $D$ ($\leq D_m$), then $j=1,2,...,D$.

In distributed ML, each worker independently computes the local gradients, and the updated gradients are aggregated to iteratively update the ML model. Given $W$ workers, in this paper, we aim to study the efficient compression of the gradients $\left\{g^{w}\right\}_{w=1}^{W}$ in order to save the communication cost among the distributed workers.

\section{The Proposed FastSGD}
\label{sec:method}
In this section, we describe the proposed FastSGD, a fast compressed SGD framework for distributed machine learning. First, we overview the FastSGD framework in Section~\ref{sec:over}. Then, we detail each component in Section~\ref{sec:gv} and Section~\ref{sec:gk}, respectively. Finally, we theoretically analyze both the space cost and the time complexity for FastSGD in Section~\ref{sec:analysis}.

\subsection{Overview of FastSGD}
\label{sec:over}
% Fig.~\ref{fig:over} illustrates the framework of the proposed FastSGD.
FastSGD consists of two stages, i.e., encoder and decoder, as described below:
\begin{itemize}
\item Encoder compresses separately the gradient values and keys  before transmitting them among the distributed workers. 
\item Decoder decompresses the compressed gradients when the gradients should be aggregated after communication.
\end{itemize}

% \begin{figure}
% \centering
% \includegraphics[width=0.43\textwidth]{overview}
% \vspace{-2mm}
% \caption{The framework of FastSGD}
% \label{fig:over}
% \end{figure}

\noindent
\textbf{The Encoder Stage of FastSGD}\quad
The encoder of FastSGD first compresses the gradient values in the following three phases.
\begin{itemize}
  \item [1)] A \textit{reciprocal mapper} is used to transform each gradient values into reciprocal values.
  \item [2)] Reciprocal gradient values are reduced to small quantized integers by a \textit{logarithm quantization}.
  \item [3)] A \textit{threshold filter} is utilized to discard the quantized integers whose absolute values are large.
\end{itemize}

%i) A \textit{reciprocal mapper} is used to transform each gradient values into reciprocal values. ii) Reciprocal gradient values are reduced to small quantized integers by a \textit{logarithm quantization}. iii) A \textit{threshold filter} is utilized to discard the quantized integers whose absolute values are large.

After compressing the gradient values, the encoder proceeds to process the corresponding gradient keys in the two phases below.
\begin{itemize}
  \item [4)] Each gradient key is represented as an incremental format \textit{delta key}.
  \item [5)] The delta keys are encoded into the binary coding, and an adaptive \textit{length flag} is further employed to save the encoding bits.
\end{itemize}

%i) Each gradient key is represented as an incremental format \textit{delta key}. ii) We encode the delta keys into the binary coding, and further employ an adaptive \textit{length flag} to save the encoding bits.

\noindent
\textbf{The Decoder Stage of FastSGD}\quad
FastSGD uses the decoder to recover the compressed gradients in the four phases as follows.
\begin{itemize}
  \item [1)] The delta keys in the binary by coding are identified using the length flags.
  \item [2)] The delta keys are recovered to the original gradient keys.
  \item [3)] The filtered integers are exponentially scaled up to reciprocal gradient values.
  \item [4)] The gradient values are recovered from the reciprocal values.
\end{itemize}
%i) The delta keys in the binary coding with length flag are recovered to the original keys. ii) The quantized values are exponentially scaled up to reciprocal gradient values. iii) The gradient values are recovered from the reciprocal values.

In what follows, we present the encoder of FastSGD, and omit the decoder of FastSGD since it simply inverses the encoder.

%Next, we describe the details of the encoder for gradient values in Section~\ref{sec:gv}, and the encoder for gradient keys in Section~\ref{sec:gk}. Theoretical analysis of FastSGD is given in Section~\ref{sec:analysis}.

%

\subsection{The Encoder for Gradient Values}
\label{sec:gv}
The prior work~\cite{DBLP:conf/sigmod/JiangFY018} has shown that the SGD optimization algorithm can converge with underestimated gradients.
Since SGD moves towards the optimal point following the opposite direction of gradients, reducing the scale of gradients might slow down the convergence rate somewhat, while still on the correct convergence track. On the contrary, the uncontrolled overestimated scale of gradients has a risk of jumping over the optimal point.

Considering the tolerance of underestimated gradients in SGD, FastSGD offers an effective three-phase gradient value encoder, which contains the reciprocal mapper, the logarithm quantization, and the threshold filter, to compress the gradient values $\{v_j\}_{j=1}^D$. We detail the three phases of gradient value encoder in the following.

\noindent
\textbf{Phase 1: Reciprocal Mapper} \quad
The reciprocal mapper transforms each gradient value $v_j$ into a reciprocal gradient value $R(v_j)$ as follows:
\begin{equation}
\label{equ:reci}
R(v_j) = \frac{\sum_{i=1}^{D} |v_i|}{|v_j|}
\end{equation}

The reciprocal mapper has two benefits/advantages. 
\begin{itemize}
  \item [1)] For each $R(v_j)$, we can get $R(v_j) \geq 1$ as $|v_j| \leq \sum_{i=1}^{D} |v_i|$. This property lays the foundation for logarithm quantization to further reduce the reciprocal value in Phase 2.
  \item [2)] A reversed gradient value order, i.e., the $v_j$ whose absolute value is bigger, would be mapped to a smaller reciprocal value. This means that we could use fewer bytes to store those important gradient elements with bigger absolute values. We would further discuss this in Phase 3.
\end{itemize}

%The first one is that for each $R(v_j)$, we can get $R(v_j) \geq 1$ as $|v_j| \leq \sum_{i=1}^{D} |v_i|$. This lays the foundation for logarithm quantization to further reduce the reciprocal value in Phase 2. The second benefit is that we can obtain a reversed value order, i.e., the $v_j$ whose absolute value is bigger would be mapped to a smaller reciprocal value. This means that we could use fewer bytes to store those important gradient elements with bigger absolute values. We would further discuss this in Phase 3.

\noindent
\textbf{Phase 2: Logarithm Quantization} \quad
After the reciprocal mapping, the logarithm quantization is utilized to further reduce the reciprocal gradient value.

Given a base $b$ ($>1$) and a reciprocal gradient value $R(v_j)$, $j=1,2,...,D$, the logarithm quantization reduces the scale of $R(v_j)$ to a smaller integer $L(v_j)$ through the following formula:
\begin{equation}
L(v_j) =  \left\lceil log_b (R(v_j)) \right\rceil
\label{eq:rgv}
\end{equation}

\begin{figure}
\centering
\includegraphics[width=0.49\textwidth]{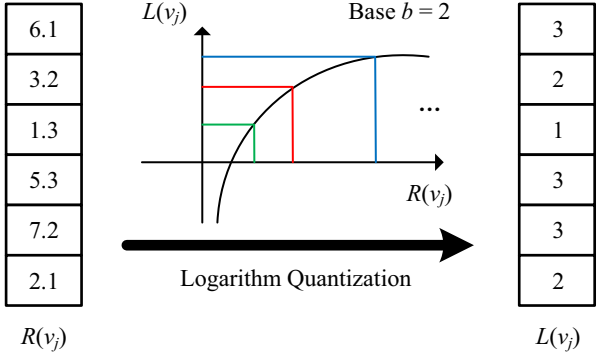}
\vspace{-2mm}
\caption{Illustration for logarithm quantization}
%\vspace{-2mm}
\label{fig:lq}
\end{figure}

Fig.~\ref{fig:lq} illustrates the process of logarithm quantization in the case $b=2$. Here, $D=6$.
Take the first reciprocal gradient value $R(v_j) = 6.1$ as an example. The logarithm quantization further reduces $R(v_j)$ into $L(v_j) =  \left\lceil log_b (R(v_j)) \right\rceil = \left\lceil log_2 6.1 \right\rceil = 3$.

Ignoring the ceiling operation in Equation (\ref{eq:rgv}), the original gradient values could be recovered precisely by $\hat{L}(v_j) = log_b (R(v_j))$ and Equation~(\ref{equ:reci}). Nonetheless, $\hat{L}(v_j)$ is decimal, and should be stored in the format of float-point number. Then, the encoding processing is meaningless because the communication cost remains the same as transmitting the original gradient value $v_j$.

Fortunately, we find the value $\hat{L}(v_j)$ has a good property that its integer part is small due to exponential explosion, yet the integer part could be a quantized indicator to estimate the original value with the base $b$. Thus, the logarithm quantization uses Equation~(\ref{eq:rgv})  with the ceiling operation to quantize the gradient value $R(v_j)$ into a small integer $L(v_j)$.

Note that, the ceiling operation is used instead of the floor operation here. This is because the ceiling operation can overestimate the quantized gradient value after the reciprocal mapper in Phase 1, which ensures the underestimate of gradient value in the decoder.

\noindent
\textbf{Phase 3: Threshold Filter} \quad
The previous two phases of gradient value encoder together complete the quantization of gradient values. Next, the threshold filter is employed to further compressed the gradient values based on sparsification ideology.

The main idea of threshold filter is based on the factor that gradient elements with larger absolute values can contribute more to the convergence of the optimization algorithm. Theoretical analysis on this has been discussed in \cite{DBLP:conf/nips/WangniWLZ18, DBLP:conf/nips/StichCJ18, DBLP:conf/nips/JiangA18}. Hence, we further filter the logarithm quantized gradient value integer with a given threshold $\tau$ in Phase 3.

After using the reciprocal mapper, we obtain a reversed gradient value order, i.e., the smaller the compressed gradient value, the bigger the absolute original gradient value. Therefore, the threshold filter discards the compressed gradient integers whose values are larger than the specified threshold $\tau$, and retains those gradient elements with larger absolute values.

After the threshold filter, the maximal value of compressed gradient integers are within the threshold $\tau$.
Hence, $\lceil log_2 \tau \rceil$ bits are sufficient to store each remaining compressed gradient value. As to be discussed in Section~\ref{sec:effect}, the hyper parameter for threshold $\tau$ is recommended to set as 128. Hence, a single byte is enough for FastSGD to encode a gradient value in practice.

So far, we have discussed the compressed gradient in the format of absolute value. However, the sign for gradient value cannot be ignored, because SGD is vulnerable to the opposite direction of gradients although it is robust to decayed gradients. Last but not the least, we assign a negative sign to the compressed gradient integers whose original gradient values are negative, and complete the encoder for gradient values.

\begin{algorithm}[t]
\caption{Gradient Value Encoder (GV-Encoder)}
\label{algo:VE}
\LinesNumbered
\DontPrintSemicolon
\KwIn{the gradient values $\{v_j\}_{j=1}^D$, a given base $b$, a given threshold $\tau$}
\KwOut{the compressed gradient values $\{L(v_j)\}_{j=1}^d$ with the $sum$ of absolute gradient values}
$sum \gets \sum_{i=1}^{D} |v_i|$ \;
$\tau_e \gets sum/{b^\tau}$ \;
$Result \gets \emptyset$ \;
\For{$j \gets 1:D$}{
    \If{$|v_j| \geq \tau_e$}{
        transform $v_j$ into the reciprocal value $R(v_j) \gets sum/|v_j|$ \;
        quantize $R(v_j)$ into the integer $L(v_j) \gets   \left\lceil log_b (R(v_j)) \right\rceil$ \;
        \If{$v_j < 0$}{
            $L(v_j) \gets -L(v_j)$ \;
        }
        $Result \gets Result \cup L(v_j)$ \;
    }
}
\Return $Result$ (i.e., $\{L(v_j)\}_{j=1}^d$) with $sum$
\end{algorithm}

Based on the above discussions, we are able to present the Gradient Value Encoder (GV-Encoder), with its pseudo-code shown in Algorithm~\ref{algo:VE}. GV-Encoder takes as inputs the gradient values $\{v_j\}_{j=1}^D$, a pre-defined base $b$, and a fixed threshold $\tau$, and outputs the compressed gradient values $\{L(v_j)\}_{j=1}^d$ with the $sum$ of absolute gradient values. Note that, the $sum$ of absolute gradient values is used for recovering the reciprocal value in the decoder, and we denote the number of filtered gradient integers as $d$ ($\leq D$). First, GV-Encoder sums up all the absolute gradient values in $sum$ (line 1). Then, it computes the early-stop threshold $\tau_e = sum/{b^\tau}$ (line 2) for filtering the gradient values before the reciprocal mapper and the logarithm quantization. Based on this, GV-Encoder avoids unnecessary processing for those two phases. Next, a set $Result$ is initialized to store the compressed gradient value integers (line 3). Thereafter, each gradient value is compressed using the three-phase encoder in a for loop (lines 4--10). Finally, GV-Encoder outputs the compressed gradient integer values $\{L(v_j)\}_{j=1}^d$ with $sum$ (line 11).

\subsection{The Encoder for Gradient Keys}
\label{sec:gk}

\begin{figure}
\centering
\includegraphics[width=0.495\textwidth]{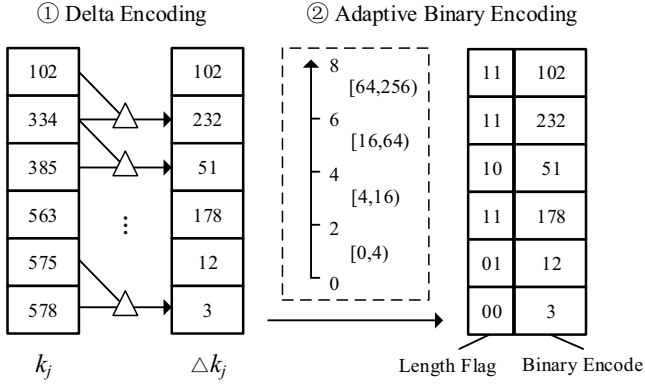}
\vspace{-2mm}
\caption{Illustration for the gradient key encoder}
%\vspace{-3mm}
\label{fig:delta}
\end{figure}

The above three-phase encoder emphasizes the compression of gradient values. In the following, we present the details of the encoder process for the sparse gradient keys $\{k_j\}_{j=1}^d$ corresponding to the compressed gradient integer values $\{L(v_j)\}_{j=1}^d$.

Different from the gradient values that could tolerate a precision loss, the gradient keys are vulnerable to errors. The recovering error in the gradient key would lead to a wrong optimal direction, and thus, the convergence of the model could not be guaranteed. Motivated by this, a lossless method that transforms the keys to delta keys has been designed~\cite{DBLP:conf/sigmod/JiangFY018}. The reason for using the delta key is that, although the original keys can be very large in many high-dimensional parameter cases, the difference between two adjacent keys is much smaller.

We follow the design of delta keys, and go a step further by using an adaptive fine-grained delta key encoding method to store gradient keys with fewer bits. Next, we detail the two following phases for gradient keys in the encoder of FastSGD.

\noindent
\textbf{Phase 4: Delta Encoding}~\cite{DBLP:conf/sigmod/JiangFY018} \quad Each gradient key $k_j$ is encoded into a delta key $\Delta k_j$, i.e., the incremental value compared with the previous key as follows:
\begin{equation}
\Delta k_j =\left\{
\begin{aligned}
& k_1, \quad \quad & j=1\\
& k_j - k_{(j-1)}, & 1<j\leq d \\
\end{aligned}
\right.
\end{equation}
\noindent
\textbf{Phase 5: Adaptive Binary Encoding} \quad After the delta encoding, the delta keys are much smaller than the original keys. In the previous work~\cite{DBLP:conf/sigmod/JiangFY018}, the byte is used to encode the delta key to replace the original format of integer or long-integer. Nonetheless, this still leaves the redundant space cost.

To alleviate this problem, we use the bit as the smallest grain to encode the delta key. Ideally, each delta key could be encoded in the binary format. For example, `3' could be binary encoded as `11' and `5' could be binary encoded as `101'. Unfortunately, the binary encoding key cannot be split to recover the original key if there is no flag to indicate the various binary encode lengths.

Towards this, we introduce the \textit{length flag} to indicate the number of bits used by each delta key. The size of length flag is denoted as $l$. In other words, we use $l$ bits to store the length flag. Then, the length flag can represent up to $2^l$ different lengths of binary encoding. By using the length flag, the delta keys could be compressed into adaptive binary encoding.

To be more specific, we take $l=2$ as an example. Assume that the maximal length of binary encoding for delta keys is $M$. Since the size $l$ of length flag is 2, the four levels of length flag, i.e., `00' to `11', could represent quaternary lengths of binary encoding, i.e., $\frac{1}{4}M$, $\frac{1}{2}M$, $\frac{3}{4}M$, and $M$, respectively. Based on this, each delta key could be encoded into a binary format that has the smallest length to store it with a length flag.

%{\color{red}Towards this, we introduce the \textit{length flag} in two bits. Assume that the maximal length of binary encoding for delta keys is $M$, the four levels of length flag, i.e., `00' to `11', could represent quaternary lengths of binary encoding, i.e., $\frac{1}{4}M$, $\frac{1}{2}M$, $\frac{3}{4}M$, and $M$, respectively. Based on this, each delta key could be encoded into a binary format that has the smallest length to store it with a length flag.}

\begin{algorithm}[t]
\caption{Gradient Key Encoder (GK-Encoder)}
\label{algo:KE}
\LinesNumbered
\DontPrintSemicolon
\KwIn{the gradient keys $\{k_j\}_{j=1}^d$}
\KwOut{the compressed gradient keys $\{\Delta k_j\}_{j=1}^d$ in binary encoding with length flag}
$preKey \gets 0$, $maxDelta \gets 0$ \;
\For{$j \gets 1:d$}{
    $\Delta k_j \gets k_j - preKey$ \;
    $preKey \gets k_j$ \;
    \If{$maxDelta < \Delta k_j$}{
        $maxDelta \gets \Delta k_j$
    }
}
$M \gets \lceil log_2 (maxDelta) \rceil$ \;
$BK \gets$ an empty binary sequence \;
\For{$j \gets 1:d$}{
    $lengthFlag \gets getLengthFlag(\Delta k_j, M)$ \;
    concatenate $lengthFlag$ with $\Delta k_j$ binary encoding to $BK$\;
}
\Return $BK$
\end{algorithm}

%Take the illustration in Fig.~\ref{fig:delta} as an example,
Fig.~\ref{fig:delta} illustrates the processing of the gradient key encoder when $l=2$. First, the gradient key encoder transforms each gradient key into a delta key. After the delta encoding, the maximum delta key is 232, which needs an 8-bit binary encoding. Hence, the quaternary ranges represented by the length flags are $[0,4)$, $[4,16)$, $[16,64)$, and $[64,256)$, respectively. Based on the length flag, each delta key could be encoded into a binary format that has the smallest length to store it with the corresponding length flag. For instance, the least key $k_j = 578$ is first encoded into the delta key $\Delta k_j = 3$. The $\Delta k_j = 3$ falls in the range of $[0,4)$, which is corresponding to the length flag '00'. Besides, it can be binary encoded as `11'. Thus, the adaptive binary encoding of the least key $k_j = 578$ is `0011'.

Algorithm~\ref{algo:KE} gives the pseudo-code of the Gradient Key Encoder (GK-Encoder). It takes the gradient key $\{k_j\}_{j=1}^d$ as input, and outputs the compressed gradient keys $\{\Delta k_j\}_{j=1}^d$ in binary encoding with length flag. First, it scans and encodes each gradient key $k_j$ into delta key $\Delta k_j$ in a for loop, and meanwhile finds the maximal delta key $maxDelta$ (lines 1--6). Then, it computes $M$ as the maximal length of binary encoding for delta keys (line 7), and initializes an empty binary sequence $BK$ to store the binary encoding with the length flag for delta keys (line 8). Next, another for loop is performed to encode each delta key by adaptive fine-grained binary encoding, and store the encoding to $BK$ in order (lines 9--11).  Finally, GK-Encoder outputs the compressed gradient keys $BK$ (line 12).

\subsection{Space Cost and Time Complexity Analysis}
\label{sec:analysis}
In this subsection, we theoretically analyze the proposed FastSGD, including both the space cost and the time complexity.

\noindent
\textbf{Space Cost of FastSGD} \quad Given a gradient consisting of key-value pairs $\{k_j, v_j\}_{j=1}^D$, FastSGD compressed it into $\{\Delta k_j, L(v_j)\}_{j=1}^d$. The size of compressed gradient integer values is $d$ bytes. For the compressed gradient delta keys, we use $D_m$ to denote the number of total parameters, and assume that the intervals of any two adjacent sparse gradient elements are consistent. Then, the expected size of each binary encoder for the delta key itself is $\lceil log_2 \frac{D_m}{d} \rceil$ bits. In addition, each length flag needs $l$ bits. Thus, the size of each compressed gradient key is $\lceil log_2 \frac{D_m}{d} \rceil + l$ bits.
As to be reported in Section~\ref{sec:effect}, the average size per gradient key is 6 bits in practice.

In summary,  the total space cost to transmit the given gradient is $8d + d(\lceil log_2 \frac{D_m}{d} \rceil + l)$ bits. In contrast, the space cost of the origin gradient is $64D$ bits even if we only use the integer format to represent the gradient key, and use the single-precision float format to store the gradient value.

\noindent
\textbf{Time Complexity of FastSGD} \quad Recall that the processes of both GV-Encoder and GK-Encoder in Section~\ref{sec:gv} and Section~\ref{sec:gk}, we could observe that the time complexity of FastSGD is $O(D)$. Thus, FastSGD achieves a fast linear time complexity.

Based on the aforementioned analysis, we can conclude that FastSGD theoretically provides an efficient compressed SGD framework with lightweight space cost for fast distributed ML. Next, we experimentally evaluate the performance of FastSGD.

\section{Experiments}
\label{sec:exp}
In this section, we conduct extensive experiments to verify the effectiveness and efficiency of the proposed FastSGD using the practical ML models and datasets, compared with the state-of-the-art competitors.

\subsection{Experimental Setup}

\begin{table}[t]
\caption{Statistics of the Datasets Used in the Experiments}
\vspace{-1mm}
\label{tab:dataset}
\begin{tabular}{|p{1.8cm}|p{2cm}|p{2cm}|p{1.3cm}|}
\hline
\textbf{Dataset} & \textbf{\# Instances} & \textbf{\# Features} & \textbf{Size (GB)} \\ \hline
\textit{URL}     & 2,396,130    & 3,231,961   & 2.1       \\ \hline
\textit{KDD10}   & 19,264,097   & 29,890,095  & 4.8       \\ \hline
\textit{KDD12}   & 149,639,105  & 54,686,452  & 20.9      \\ \hline
\textit{WebSpam} & 350,000       & 16,609,143  & 23.3      \\ \hline
\end{tabular}
\end{table}

\noindent
\textbf{Datasets} \quad We employ four public datasets, namely, \textit{URL}, \textit{KDD10}, \textit{KDD12}, and \textit{WebSpam}, with their statistics listed in Table~\ref{tab:dataset}.
\textit{URL} is commonly used to train machine learning models for malicious URL detection{\footnote{http://www.sysnet.ucsd.edu/projects/url/}}, and consists of 2 million instances with 3 million features.
\textit{KDD10} is a public dataset published by KDD Cup 2010{\footnote{https://pslcdatashop.web.cmu.edu/KDDCup/}}, including 19 million instances and 29 million features. The task is to predict student performance on mathematical problems from logs of student interaction with Intelligent Tutoring Systems.
\textit{KDD12} is published by another generation KDD Cup{\footnote{https://www.kaggle.com/c/kddcup2012-track1}}, which contains 149 million instances and 54 million features. The challenge involves predicting whether or not a user will follow an item that has been recommended to the user.
\textit{WebSpam} is a publicly available Web spam dataset{\footnote{https://www.cc.gatech.edu/projects/doi/WebbSpamCorpus.html}}. The corpus consists of nearly 350,000 Web spam pages with 16 million features.

\noindent
\textbf{ML models} \quad To evaluate the performance of FastSGD, we choose three popular ML methods, as with \cite{DBLP:conf/sigmod/JiangFY018}, including Linear Regression (Linear), Logistic Regression (LR), and Support Vector Machine (SVM).
% Their formalized loss functions are listed below:
% \begin{equation}
% \begin{array}{l}
% Linear: f(x, y, \theta)=\sum_{i=1}^{N}\left(y_{i}-\theta^{\mathrm{T}} x_{i}\right)^{2}+\frac{\lambda}{2}\|\theta\|_{2} \\
% LR: f(x, y, \theta)=\sum_{i=1}^{N} \log \left(1+e^{-y_{i} \theta^{\mathrm{T}} x_{i}}\right)+\frac{\lambda}{2}\|\theta\|_{2}\\
% SVM: f(x, y, \theta)=\sum_{i=1}^{N} \max \left(0,1-y_{i} \theta^{\mathrm{T}} x_{i}\right)+\frac{\lambda}{2}\|\theta\|_{2}
% \end{array}
% \end{equation}
Their formalized descriptions are listed in Table~\ref{tab:ml}, in which $\mathbb{I}$ represents the indicator function.

\begin{table*}[t]
\caption{Formalization of the Machine Learning Models Evaluated}
\label{tab:ml}
\vspace{-1mm}
\begin{tabular}{|p{4.2cm}|p{6.3cm}|p{6.3cm}|}
\hline
\textbf{Machine learning model} & \textbf{Loss function} & \textbf{Gradient}  \\ \hline
Linear regression (Linear) & $ f(x, y, \theta)=\sum_{i=1}^{N}\left(y_{i}-\theta^{\mathrm{T}} x_{i}\right)^{2}+\frac{\lambda}{2}\|\theta\|_{2}$ & $ \nabla_\theta f(x,y,\theta) = \sum_{i=1}^{N}-\left(y_{i}-\theta^{\mathrm{T}} x_{i}\right) x_{i}+\lambda \theta$\\  \hline
Logistic regression (LR)  & $f(x, y, \theta)=\sum_{i=1}^{N} \log \left(1+e^{-y_{i} \theta^{\mathrm{T}} x_{i}}\right)+\frac{\lambda}{2}\|\theta\|_{2}$  & $\nabla_\theta f(x,y,\theta) = \sum_{i=1}^{N}-\frac{y_{i}}{1+ e^{\left(y_{i} \theta^{\mathrm{T}} x_{i}\right)}} x_{i}+\lambda \theta$   \\ \hline
Support vector machine (SVM) & $f(x, y, \theta)=\sum_{i=1}^{N} \max \left(0,1-y_{i} \theta^{\mathrm{T}} x_{i}\right)+\frac{\lambda}{2}\|\theta\|_{2}$ & $ \nabla_\theta f(x,y,\theta) =\sum_{i=1}^{N}-y_{i} x_{i} \mathbb{I}\left\{y_{i} \theta^{\mathrm{T}} x_{i}<1\right\}+\lambda \theta$\\ \hline
\end{tabular}
\end{table*}

To converge faster, we train three ML models with Adam SGD~\cite{DBLP:journals/corr/KingmaB14}. Adam SGD is one of the most commonly-used accelerated versions of SGD. At the iterative round $t+1$, Adam SGD stores a decaying average of past gradients and squared gradients:
\begin{equation}
\begin{array}{l}
m_{t}=\beta_{1} m_{t-1}+\left(1-\beta_{1}\right) g_{t} \\
v_{t}=\beta_{2} v_{t-1}+\left(1-\beta_{2}\right) g_{t}^{2}
\end{array}
\end{equation}
where $\beta_{1}$ and $\beta_{2}$ are two hyper-parameters close to 1. Then, Adam SGD updates the model parameters as follows:
\begin{equation}
\theta_{t+1}=\theta_{t}-\frac{\eta}{\sqrt{v_{t}+\epsilon}} m_{t}
\end{equation}
where $\epsilon$ is another hyper-parameters close to 0.

\noindent
\textbf{Implementation} \quad
Following the choice of \cite{DBLP:conf/sigmod/JiangFY018}, we set  $\beta_{1}$ to 0.9, $\beta_{2}$ to 0.999, and $\epsilon$ to $10^{-8}$  in Adam SGD. The optimal learning rate $\eta$ is tuned by a grid search. The regularization hyper-parameter $\lambda$ is set to 0.01. The input dataset is split into the training dataset (70\% of the input dataset) and the test dataset (30\% of the input dataset).
To decrease the synchronization frequency and the communication cost, we implement SGD with a popular \textit{mini-batch} trick~\cite{DBLP:conf/compstat/Bottou10}, which uses a batch of (more than one) instances in each training iteration.
We set the batch size as 10\% of the training dataset. Once SGD completes a pass over the entire training dataset, we say SGD has finished an epoch. We verify the performance of SGD in 20 epochs by default without special instructions.

\noindent
\textbf{Competitors} \quad We compare FastSGD with four state-of-the-art algorithms, including Adam SGD~\cite{DBLP:journals/corr/KingmaB14}, LogQuant~\cite{DBLP:journals/corr/MiyashitaLM16}, SketchML~\cite{DBLP:conf/sigmod/JiangFY018}, and Top-$k$~\cite{DBLP:conf/ijcai/ShiZWTC19}.

\begin{itemize}
\item{} Adam SGD is one of the most popular first-order gradient optimization approaches, which combines the advantage of momentum~\cite{DBLP:journals/nn/Qian99} and adaptive learning rate~\cite{DBLP:journals/jmlr/DuchiHS11}.
\item{} LogQuant adopts base-2 logarithmic representation to quantize gradients.
\item{} SketchML is a sketched-based method to compress the sparse gradient consisting of key-value pairs.
\item{} Top-$k$ selects the $k$ largest values in the gradient to transmit, which can achieve good performance in practice.
\end{itemize}

Note that the Adam strategy is applied to all the competitors for fairness.

The main performance metrics to study the gradient compression performance include the average running time per epoch, the message size, and the loss function with respect to the running time.

All the experiments{\footnote{We sincerely thank the authors of SketchML~\cite{DBLP:conf/sigmod/JiangFY018} for sharing the source code.}} were implemented in Scala 2.11 on Spark 2.2, and run on a 12-
node Dell cluster. Each node has two Intel(R) Xeon(R) CPU E5-2650 v4 @ 2.20GHz processors with 12 cores, 128GB RAM, 1TB disk, and connect in Gigabit Ethernet.

\subsection{Effectiveness of the Proposed Techniques}
\label{sec:effect}

In this subsection, we experimentally evaluate the effectiveness of the techniques proposed in this paper on \textit{KDD12} dataset. Note that, the results over the three other datasets are similar.

\begin{figure}[t]
\centering
\includegraphics[width=0.45\textwidth]{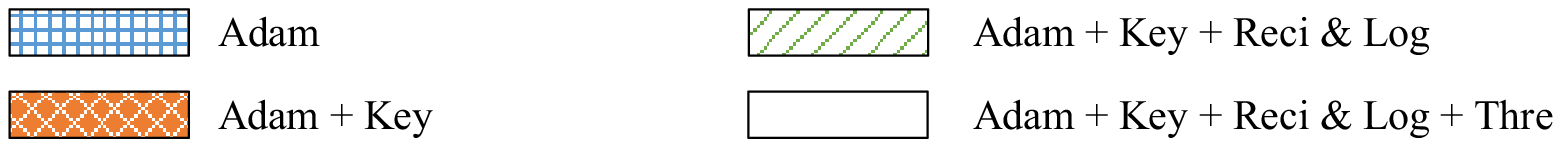}\\
%\vspace{-3mm}
  \subfigure[running time per epoch]{
   \label{fig:component-time}
   \includegraphics[width=0.49\textwidth]{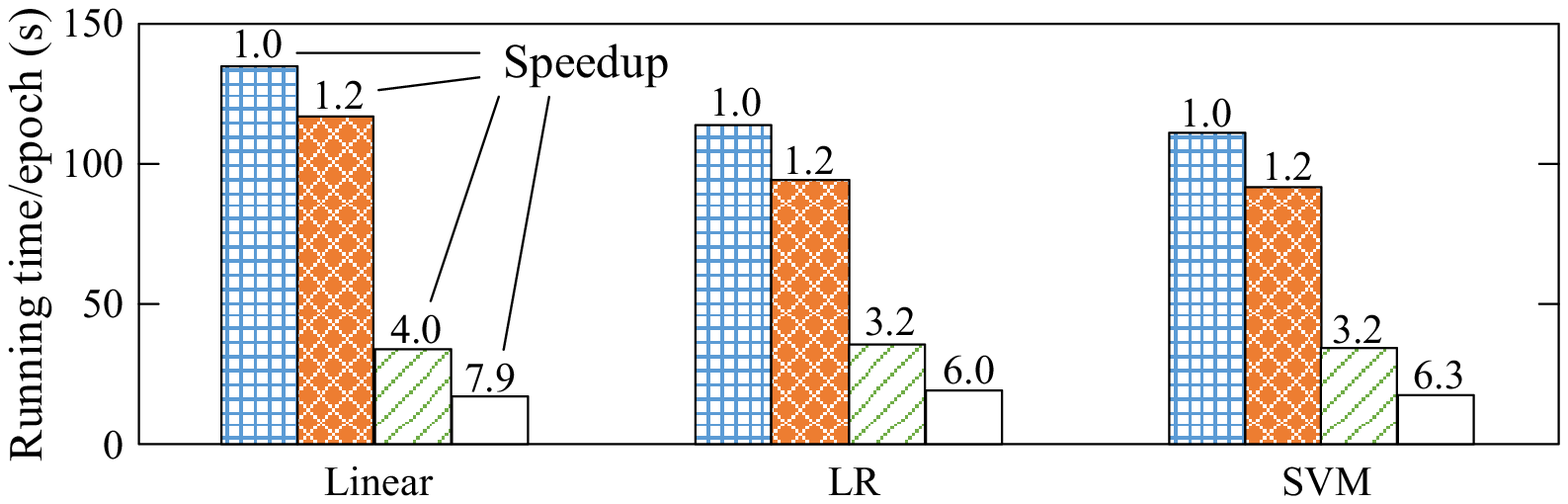}
  }
  \subfigure[message size]{
   \label{fig:component-size}
   \includegraphics[width=0.49\textwidth]{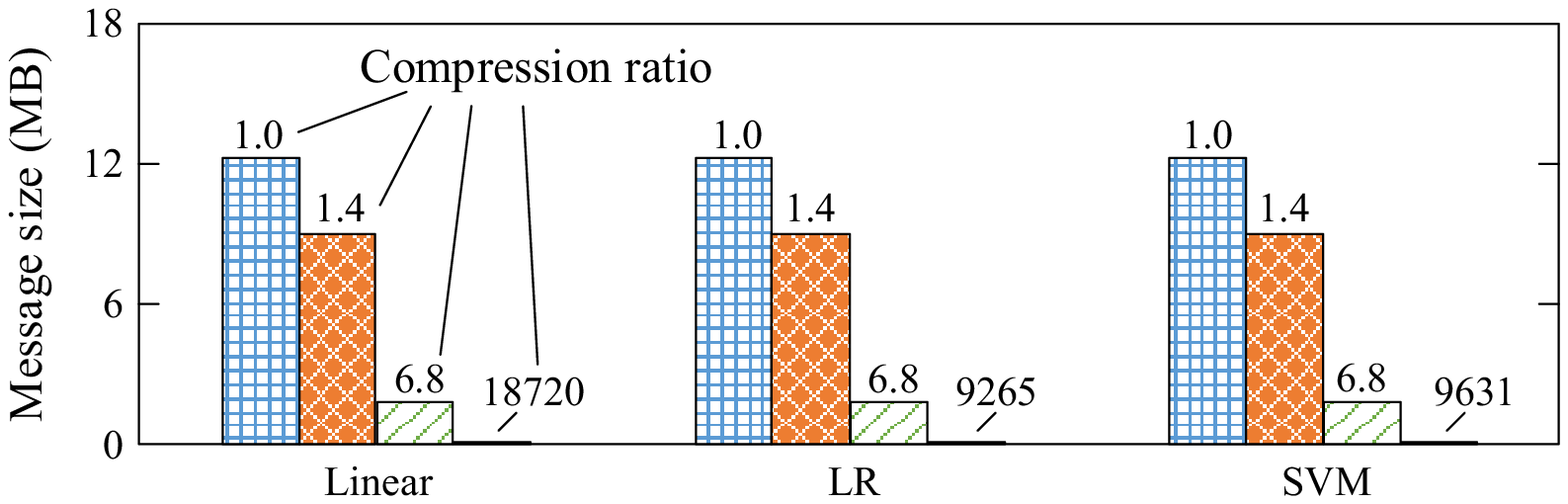}
  }
\vspace{-2mm}
\caption{Effectiveness of the proposed components on \textit{KDD12}}
\label{fig:component}
\end{figure}

\noindent
\textbf{Effect of the Proposed Components} \quad FastSGD could be divided into three components, i.e., i) the adaptive binary encoder for delta gradient key, ii) the reciprocal mapper and logarithm quantization that map decimal gradient values into small quantized integers, and iii) the threshold filter to discard the quantized gradient value. We begin with the baseline Adam, and implement the proposed components gradually. Fig.~\ref{fig:component} illustrates the performance of the proposed components.

Fig.~\ref{fig:component-time} plots the running time per epoch with the speedup compared with the basic Adam in three different ML models. It is observed that the proposed techniques can significantly improve the efficiency of three different ML models. Compared with Adam, the component of adaptive binary encoder accelerates the execution by 1.2$\times$. In addition, the reciprocal mapper and logarithm quantization further improves the performance by up to 4.0$\times$. Finally, FastSGD achieves the speedup for the running time per epoch at most 7.9$\times$ by implementing with all the three components.

The reason for the speedup is that the proposed components compress the communication volume. Fig.~\ref{fig:component-size} shows the message size in each round of communication (i.e., per batch) with the compression ratio compared against the uncompressed message size. FastSGD achieves the compression ratio from 1.4$\times$ to 18720$\times$ at most by consolidating three components. We can observe that the threshold filter remarkably compresses the message since it discards lots of gradients whose absolute values are small. Nevertheless, the running time is not decreased in the same proportion as the message size. This is because FastSGD achieves a high compression ratio for the gradient communication volume, which makes the computation ability rather than the communication become the bottleneck of the distributed system.

\begin{table*}[ht]
\caption{Time Breakdown on \textit{KDD12}}
\vspace{-1mm}
\label{tab:break}
\center
\begin{tabular}{|p{2cm}|l|l|l|l|l|l|}
\hline
\multicolumn{7}{|c|}{\textbf{Linear}}                                                                                                                                          \\ \hline
\textbf{Method} & \textbf{Total time} & \textbf{Gradient computation} & \textbf{Gradient encoder} & \textbf{Gradient decoder} & \textbf{Communication} & \textbf{Model updating} \\ \hline
Adam            & 186                 & 22                            & 0                         & 0                         & 84                     & 80                    \\ \hline
FastSGD         & 33                  & 21                            & 0.05                      & 0.08                      & 6                      & 6                     \\ \hline
\multicolumn{7}{|c|}{\textbf{LR}}                                                                                                                                              \\ \hline
\textbf{Method} & \textbf{Total time} & \textbf{Gradient computation} & \textbf{Gradient encoder} & \textbf{Gradient decoder} & \textbf{Communication} & \textbf{Model updating} \\ \hline
Adam            & 156                 & 23                            & 0                         & 0                         & 68                     & 65                    \\ \hline
FastSGD         & 32                  & 23                            & 0.05                      & 0.06                      & 4                      & 5                     \\ \hline
\multicolumn{7}{|c|}{\textbf{SVM}}                                                                                                                                              \\ \hline
\textbf{Method} & \textbf{Total time} & \textbf{Gradient computation} & \textbf{Gradient encoder} & \textbf{Gradient decoder} & \textbf{Communication} & \textbf{Model updating} \\ \hline
Adam            & 158                 & 21                            & 0                         & 0                         & 73                     & 64                    \\ \hline
FastSGD         & 34                  & 21                            & 0.06                        & 0.07                        & 5                      & 8                     \\ \hline
\end{tabular}
\end{table*}

Furthermore, Table~\ref{tab:break} reports a breakdown of the running time per epoch in seconds to depict the raw performance of the encoder and decoder in FastSGD. We decouple the total running time into five parts, i.e., gradient computation, gradient encoding, gradient decoding, communication, and model updating. It is observed that the encoder and decoder in FastSGD spend a small amount of time (i.e., 0.05 to 0.08 seconds) to accomplish the gradient compression and decompression efficiently. Based on this, FastSGD achieves up to 17$\times$ speedup in the communication time and 13$\times$ speedup in the model updating time. The reason for the speedup in the communication time is that FastSGD efficiently compresses the gradient communication volume. Besides, FastSGD accelerates the model updating due to the threshold filter, which discards those gradient elements with smaller absolute values and hence reduces the model updating computation cost.

\begin{table}[t]
\caption{The Space Cost of Gradient Key on \textit{KDD12}}
\vspace{-1mm}
\label{tab:key}
\begin{tabular}{|p{2.6cm}|p{1.5cm}|p{1.5cm}|p{1.5cm}|}
\hline
\textbf{Method}  & \textbf{Adam} & \textbf{SketchML} & \textbf{FastSGD} \\ \hline
Size per key (bits) & 32.00    & 9.95     & 6.04    \\ \hline
\end{tabular}
\end{table}

As discussed in Section~\ref{sec:gk}, both SketchML and FastSGD use the delta gradient key. Further, FastSGD utilizes the bit rather than byte as the smallest grain with the adaptive strategy to encode the delta key. Thus, we record the average size taken by each gradient key in Table~\ref{tab:key} to verify the effectiveness of FastSGD. It is observed that SketchML compresses the average size per gradient key from 32.00 bits to 9.95 bits, while FastSGD further reduces it to 6.04 bits. This confirms the effectiveness of adaptive fine-grained delta encoding method in FastSGD.

\noindent
\textbf{Effect of the Hyper-parameters in FastSGD} \quad FastSGD contains three hyper-parameters, i.e., the base $b$ in logarithm quantization, the threshold $\tau$ in threshold filter, and the size $l$ of length flag. We fine-tune them, and give the setting guidance based on the experimental results over all the three different ML models.

%\textbf{Effect of the hyper-parameters in FastSGD} \quad FastSGD contains two hyper-parameters, i.e., the base $b$ in logarithm quantization and the threshold $\tau$ in threshold filter. We fine-tune them and give the setting guidance based on the experimental results over Linear and LR models.

\begin{figure*}[t]
\centering
\includegraphics[width=0.7\textwidth]{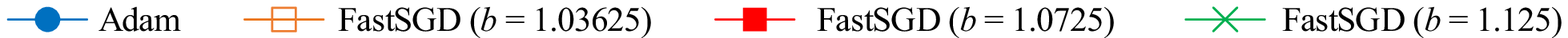}\\
\vspace{-1mm}
  \subfigure[Linear]{
   \label{base-Linear}
   \includegraphics[width=0.235\textwidth]{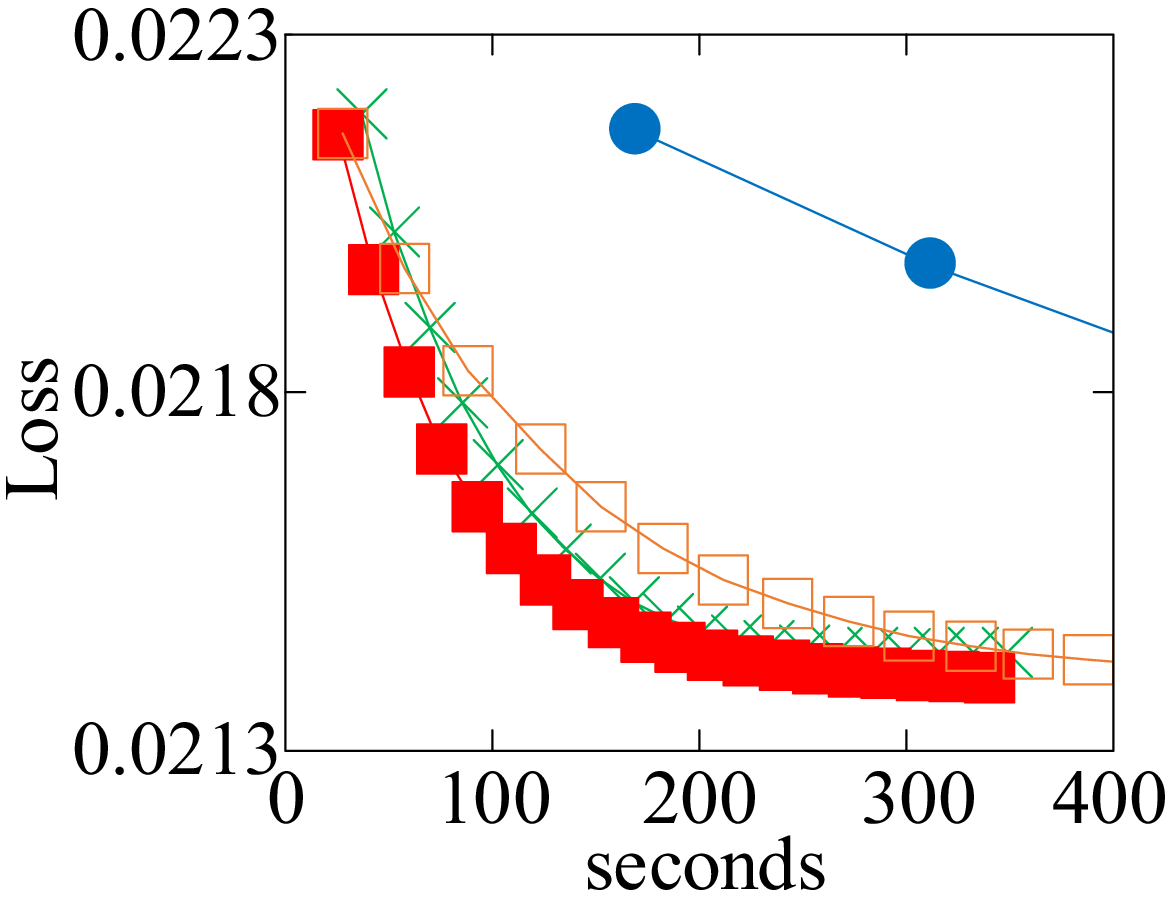}
  }
  \hspace{10mm}
  \subfigure[LR]{
   \label{base-LR}
   \includegraphics[width=0.235\textwidth]{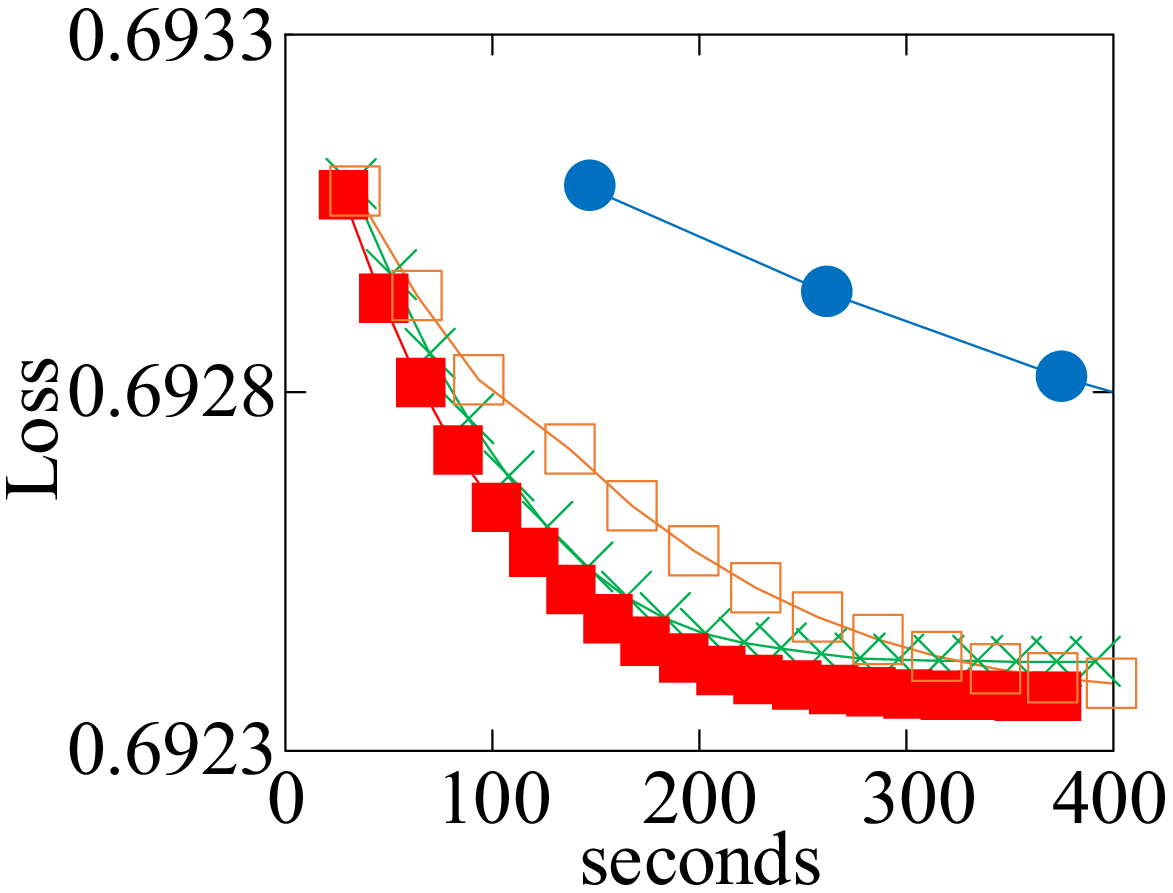}
  }
  \hspace{10mm}
  \subfigure[SVM]{
   \label{base-SVM}
   \includegraphics[width=0.235\textwidth]{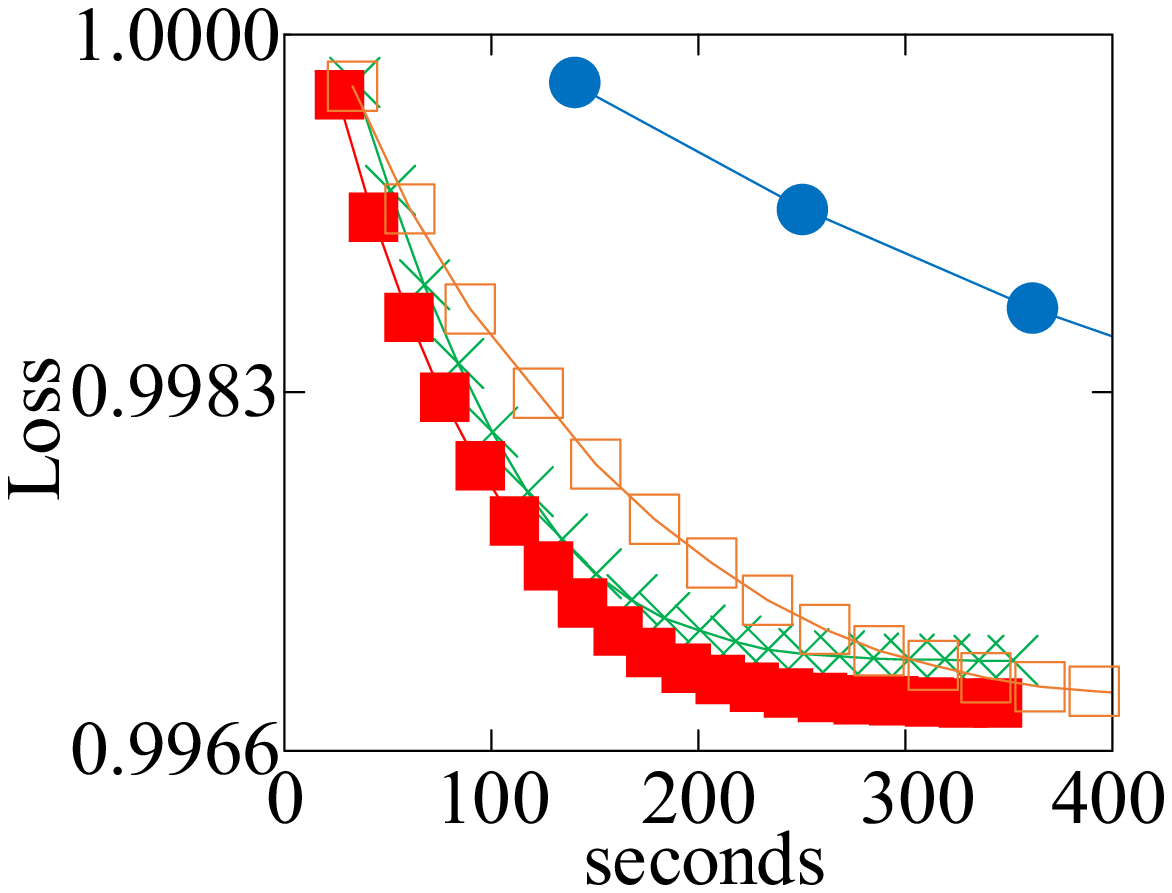}
  }
\vspace{-2mm}
\caption{Effect of base $b$ on \textit{KDD12}}
\label{fig:base}
\end{figure*}

\begin{figure*}[t]
\centering
\includegraphics[width=0.7\textwidth]{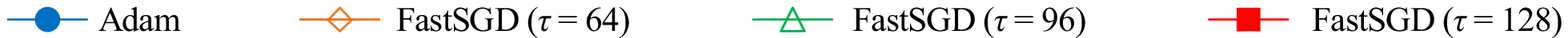}\\
\vspace{-1mm}
  \subfigure[Linear]{
   \label{thre-Linear}
   \includegraphics[width=0.235\textwidth]{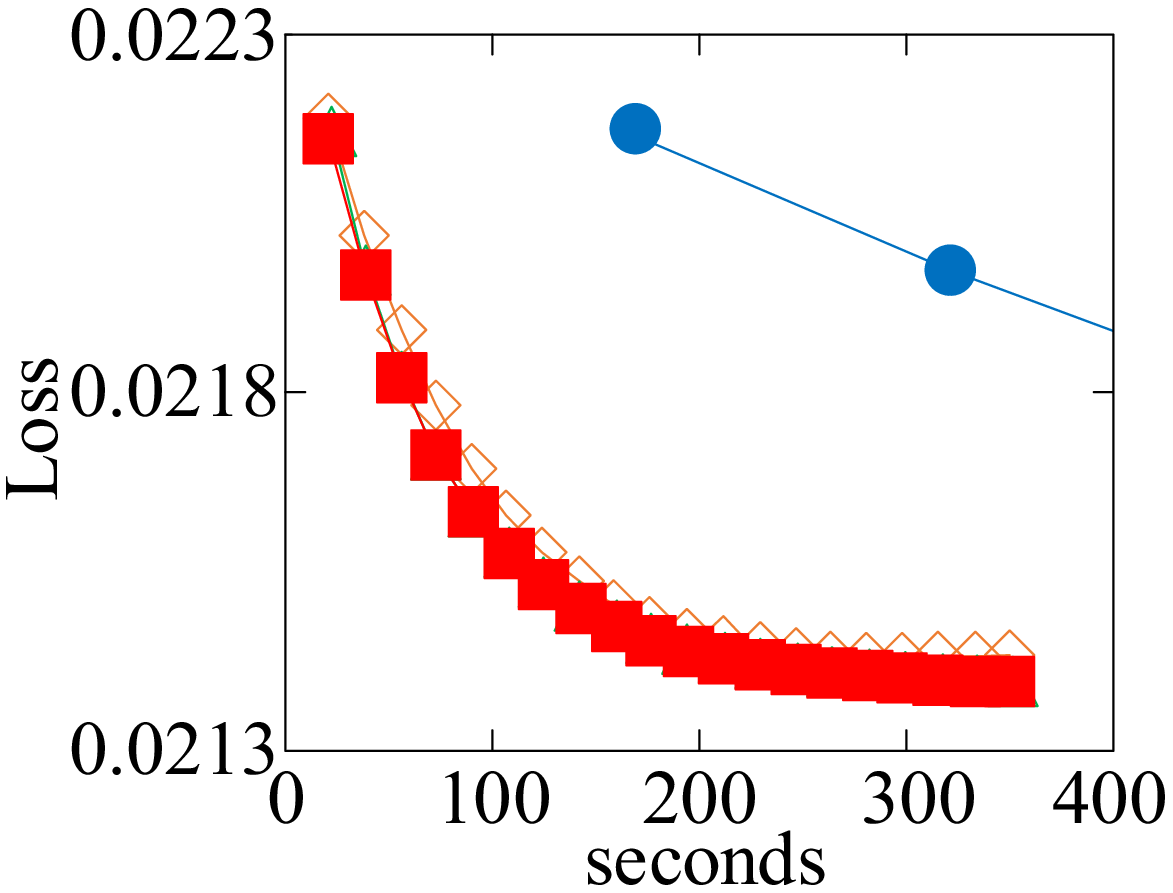}
  }
  \hspace{10mm}
  \subfigure[LR]{
   \label{thre-LR}
   \includegraphics[width=0.235\textwidth]{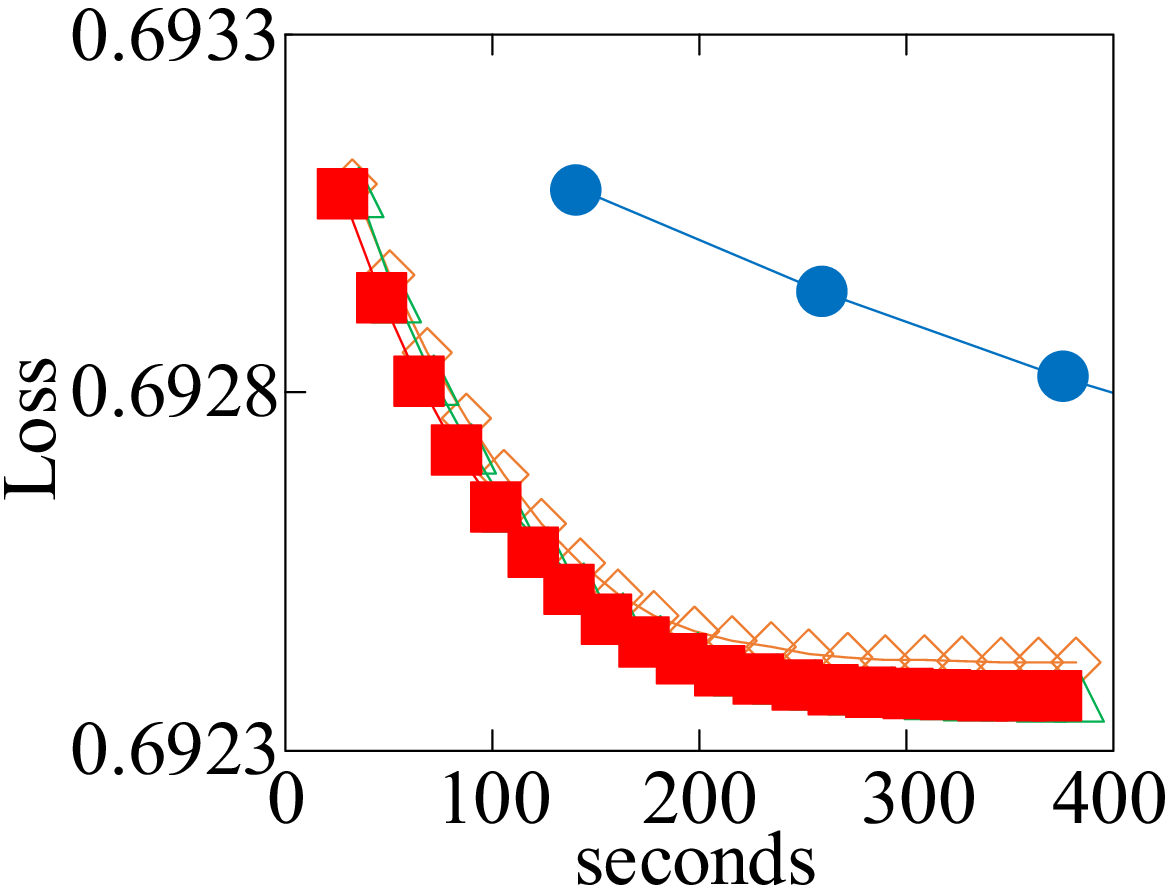}
  }
  \hspace{10mm}
  \subfigure[SVM]{
   \label{thre-SVM}
   \includegraphics[width=0.235\textwidth]{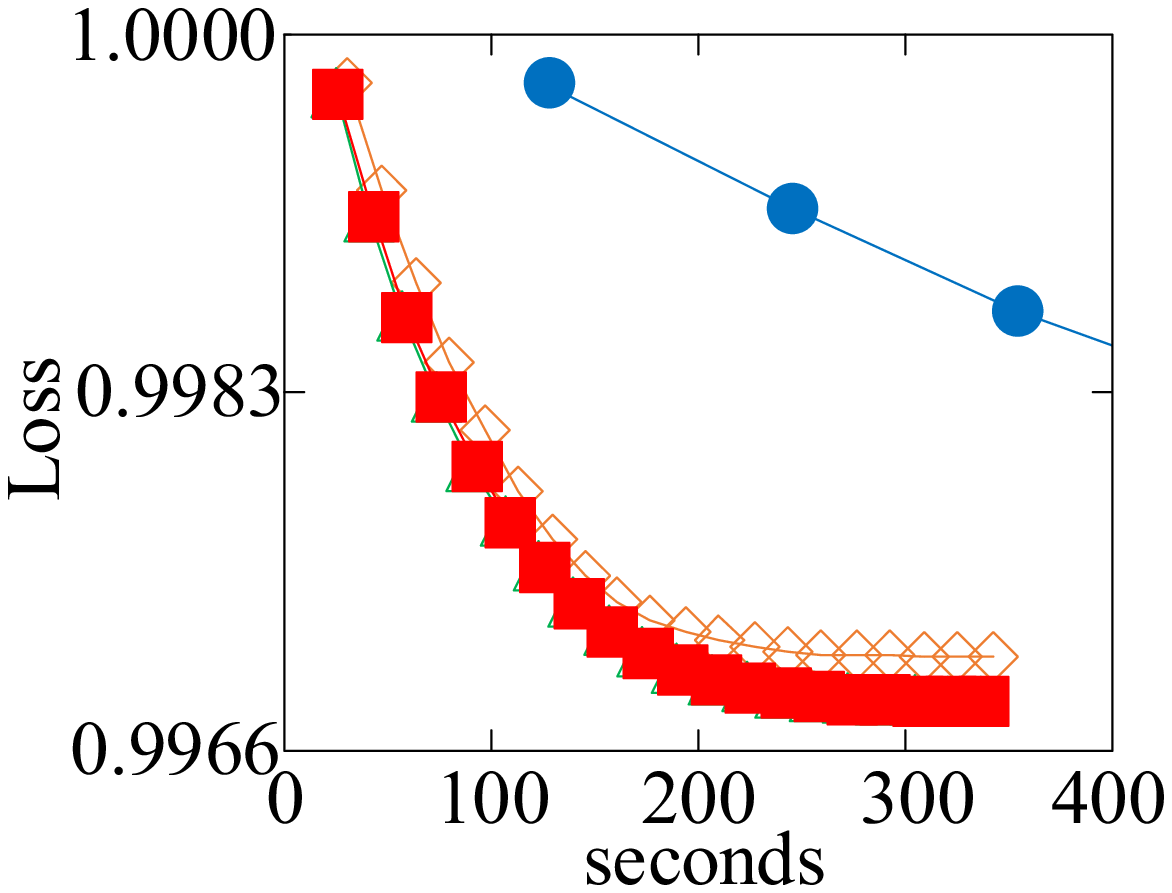}
  }
\vspace{-2mm}
\caption{Effect of threshold $\tau$ on \textit{KDD12}}
\label{fig:thre}
\end{figure*}

Fig.~\ref{fig:base} depicts the loss function of each epoch in terms of the running time with the base $b$ changing from 1.03625 to 1.1125. As excepted, FastSGD converges much faster than the baseline Adam in all three base values. Besides, the convergence rate of FastSGD first becomes faster and then slower as the base $b$ grows, because the bigger base $b$, the smaller the gradient values are compressed, while the recovering error in decoder becomes larger. To achieve a good trade-off between the compression power and the decompression error, the value of base $b$ is recommended to set as around 1.1.

Fig.~\ref{fig:thre} shows the loss function of each epoch in terms of the running time w.r.t. the threshold $\tau$ varying from 64 to 128. We observe that the results of FastSGD on different thresholds are similar. A bigger threshold $\tau$ (i.e., 128), which retains more gradient values and restricts the space cost for each gradient value in one byte, could perform slightly better. Therefore, we set the threshold $\tau$ to 128 by default in our experiments.

Table~\ref{tab:vkey} lists the average size per gradient key varying the size $l$ of length flag from 1 to 5 bits. It is observed that the average size per gradient key first becomes smaller and then larger as the size $l$ of length flag grows. The reason is that the larger length flag provides more fine-grained length options to store the delta key in less bits, while it costs more space to represent the length flag itself. According to the experimental results, the size $l$ of length flag is recommended to set as 2.

\begin{table}[t]
\caption{The Space Cost of Gradient Key v.s. the Size of Length Flag on \textit{KDD12}}
\vspace{-1mm}
\label{tab:vkey}
\begin{tabular}{|p{2.6cm}|p{0.72cm}|p{0.72cm}|p{0.72cm}|p{0.72cm}|p{0.72cm}|}
\hline
\textbf{Size of length flag} & \textbf{1} & \textbf{2} & \textbf{3} & \textbf{4} & \textbf{5} \\ \hline
Size per key (bits) &  8.65 & 6.04 & 7.06 &  12.27 & 14.78  \\ \hline
\end{tabular}
\end{table}

\begin{figure*}[ht]
\centering
  \subfigure[Linear]{
   \label{feature-Linear}
   \includegraphics[width=0.235\textwidth]{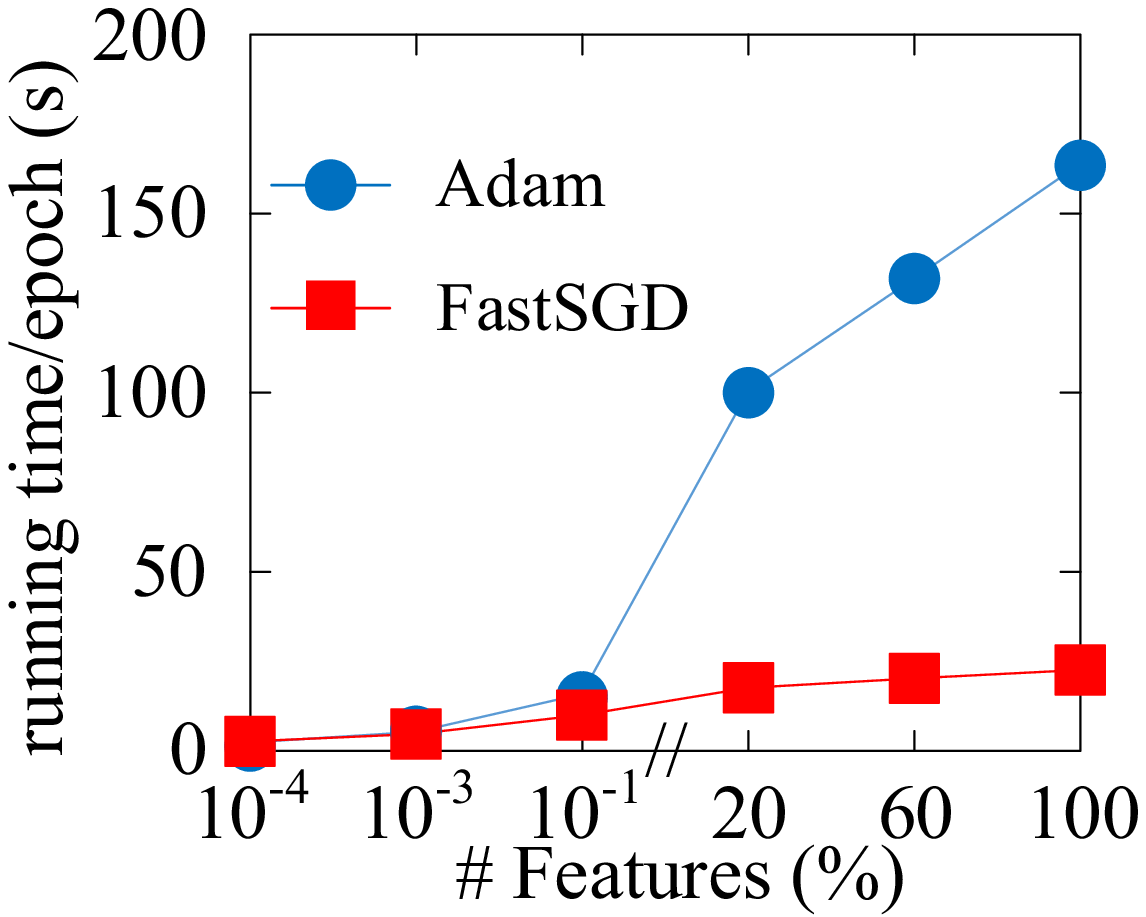}
  }
  \hspace{10mm}
  \subfigure[LR]{
   \label{feature-LR}
   \includegraphics[width=0.235\textwidth]{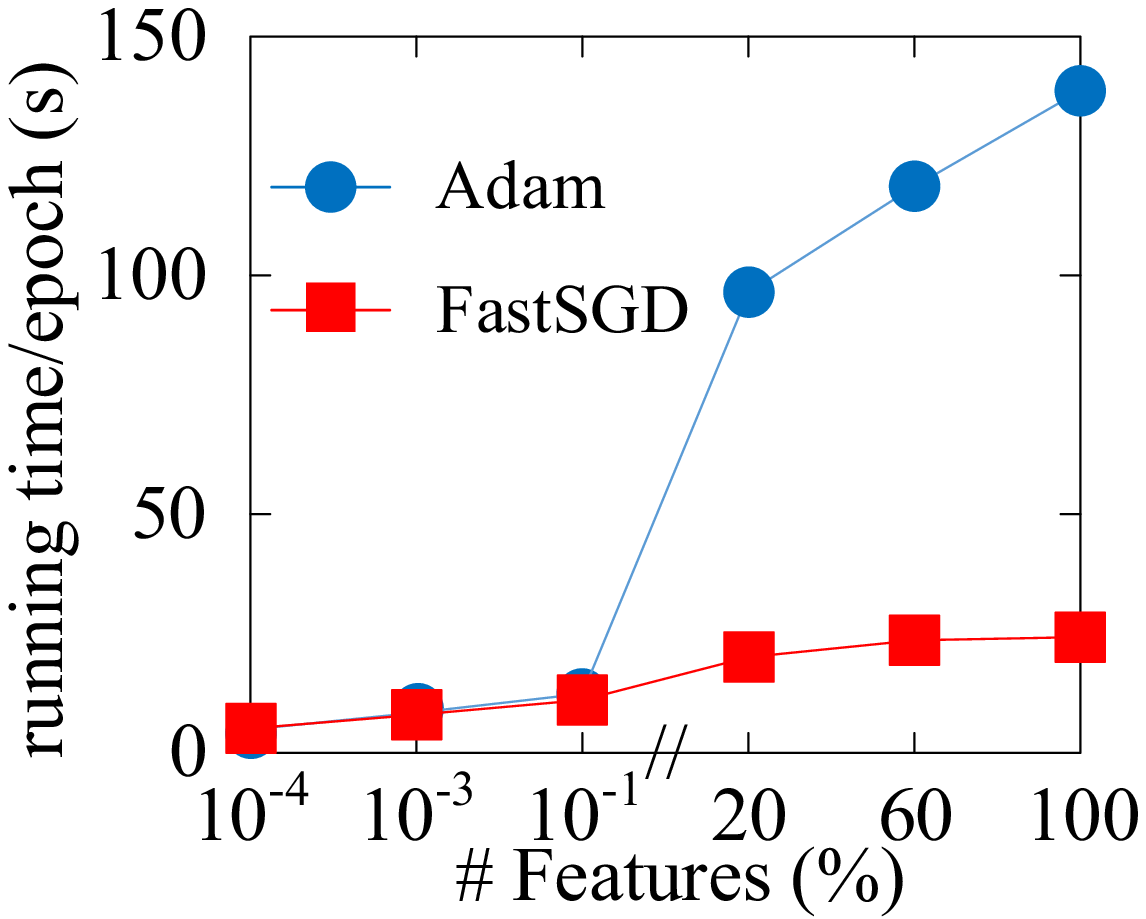}
  }
  \hspace{10mm}
  \subfigure[SVM]{
   \label{feature-SVM}
   \includegraphics[width=0.235\textwidth]{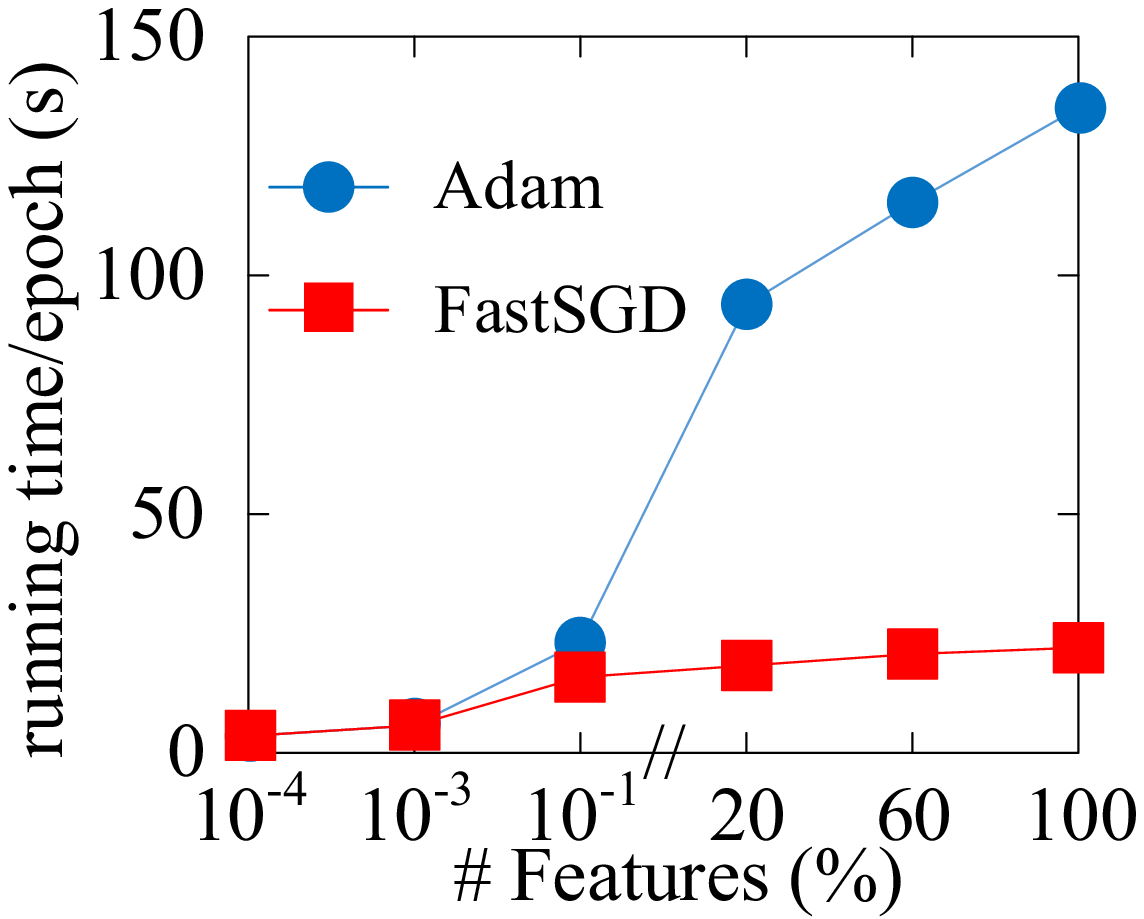}
  }
\vspace{-2mm}
\caption{Effect of the number of features on \textit{KDD12}}
\label{fig:feature}
\end{figure*}

\noindent
\textbf{Effect of the Number of Features} \quad In this set of experiments, we explore the scalability of FastSGD with fewer features. Fig.~\ref{fig:feature} depicts the running time per epoch when the number of features changes from $10^{-4}$\% to 100\% of the whole \textit{KDD12} dataset. As expected, the running time per epoch of both FastSGD and Adam increases as the number of features ascends. This is because there are more feature data to be processed. In addition, we can observe that FastSGD outperforms Adam in almost all the experimental settings, except when the number of features is extremely small (e.g., $10^{-4}$\% of the whole feature number). The reason is that the cost for gradient compression exceeds the savings by the reduced gradient communication with such a small number of features.

\subsection{Comparison with the State-of-the-arts}

In this subsection, we compare the end-to-end performance of our proposed FastSGD with four state-of-the-art algorithms, viz., Adam, LogQuant, SketchML, and Top-$k$.

\begin{figure*}[t]
\centering
\includegraphics[width=0.95\textwidth]{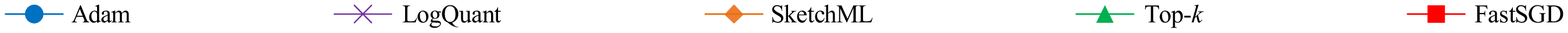}\\
\vspace{-1mm}
  \subfigure[\textit{URL}]{
   \label{fig:competitor-URL}
   \includegraphics[width=0.235\textwidth]{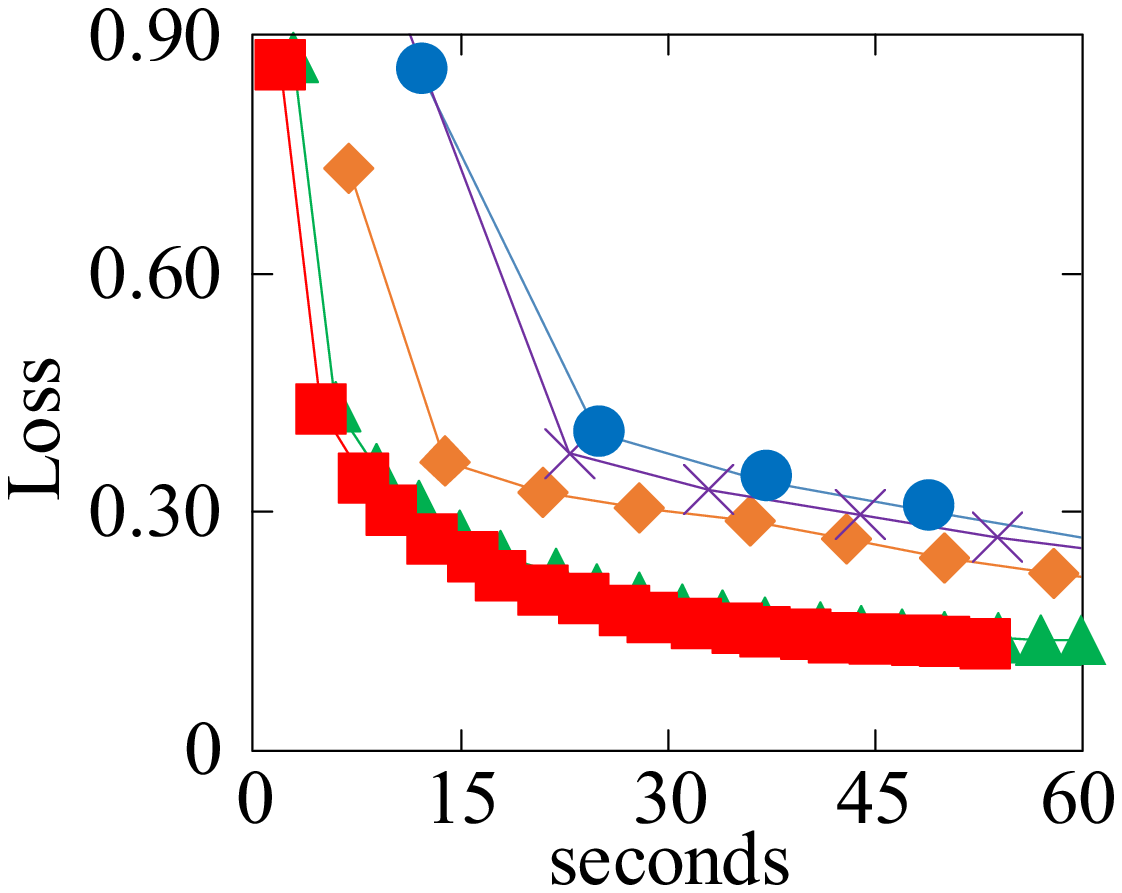}
  }%\vspace{-3mm}
  \subfigure[\textit{KDD10}]{
   \label{fig:competitor-KDD10}
   \includegraphics[width=0.235\textwidth]{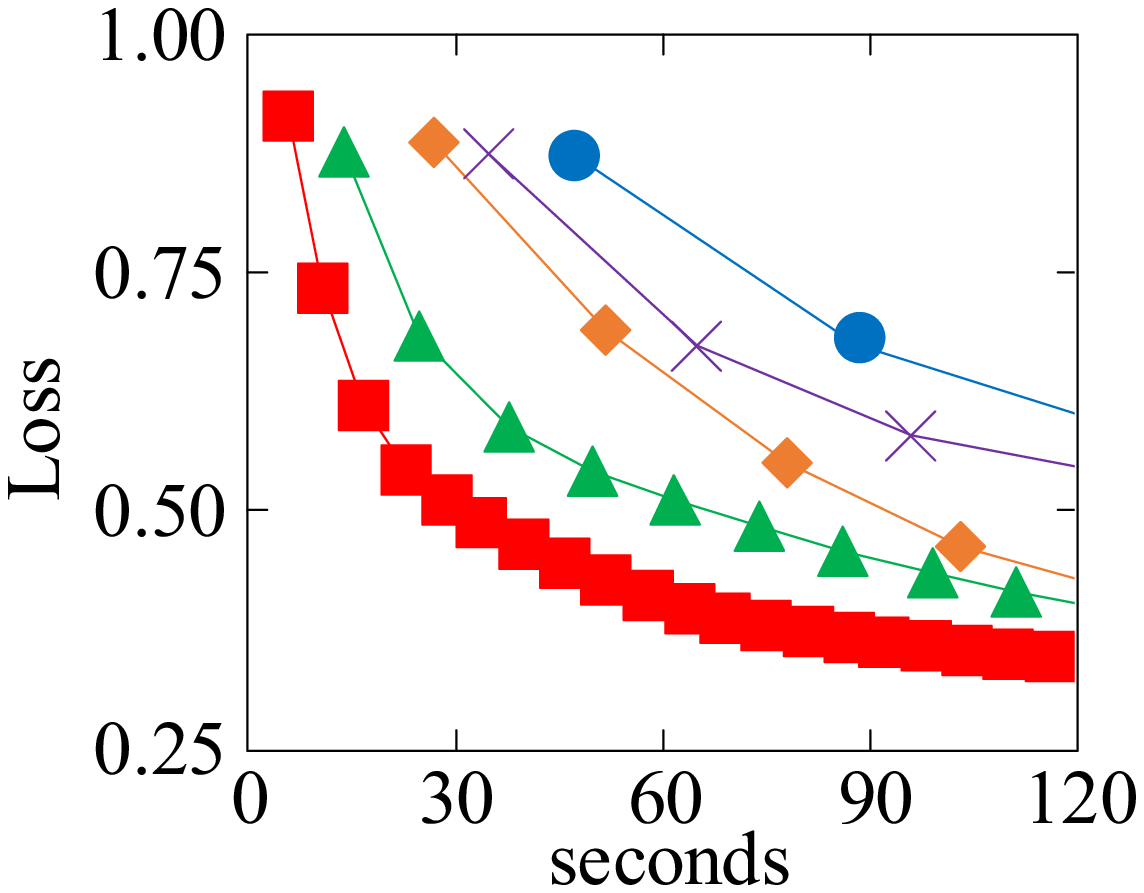}
  }
  \subfigure[\textit{KDD12}]{
   \label{fig:competitor-KDD12}
   \includegraphics[width=0.235\textwidth]{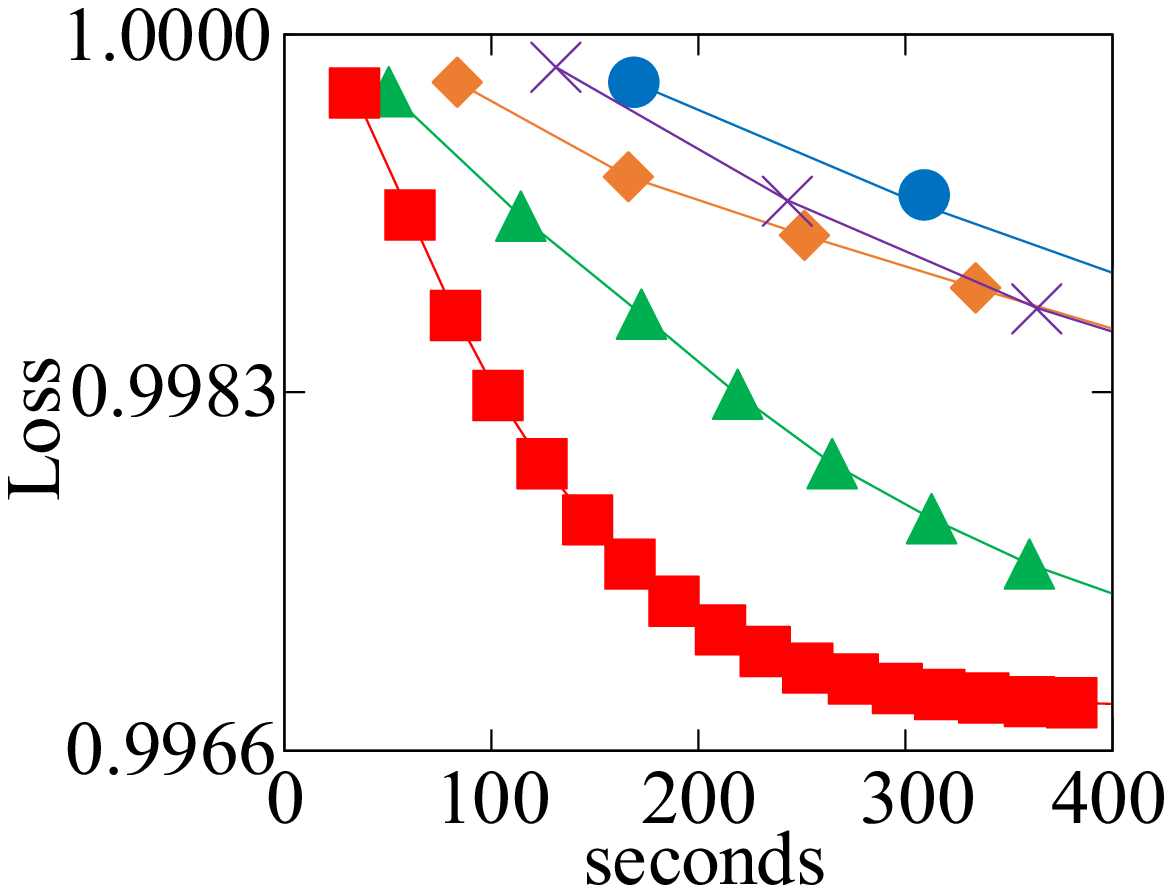}
  }
  \subfigure[\textit{WebSpam}]{
   \label{fig:competitor-Web}
   \includegraphics[width=0.235\textwidth]{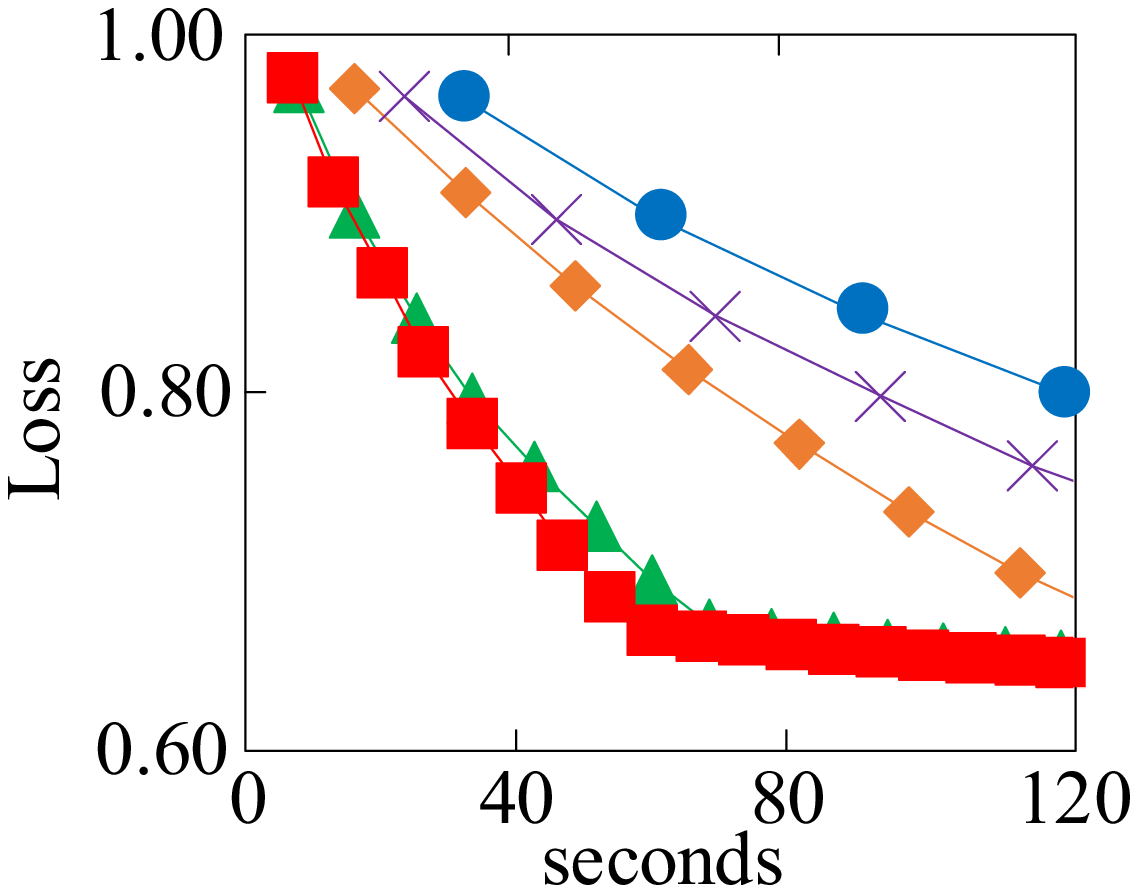}
  }
\vspace{-2mm}
\caption{Convergence rate over SVM}
\label{fig:competitor}
\end{figure*}

\begin{figure*}[t]
\centering
\includegraphics[width=0.95\textwidth]{competitor-icon}\\
\vspace{-1mm}
  \subfigure[\textit{URL}]{
   \label{fig:worker-URL}
   \includegraphics[width=0.235\textwidth]{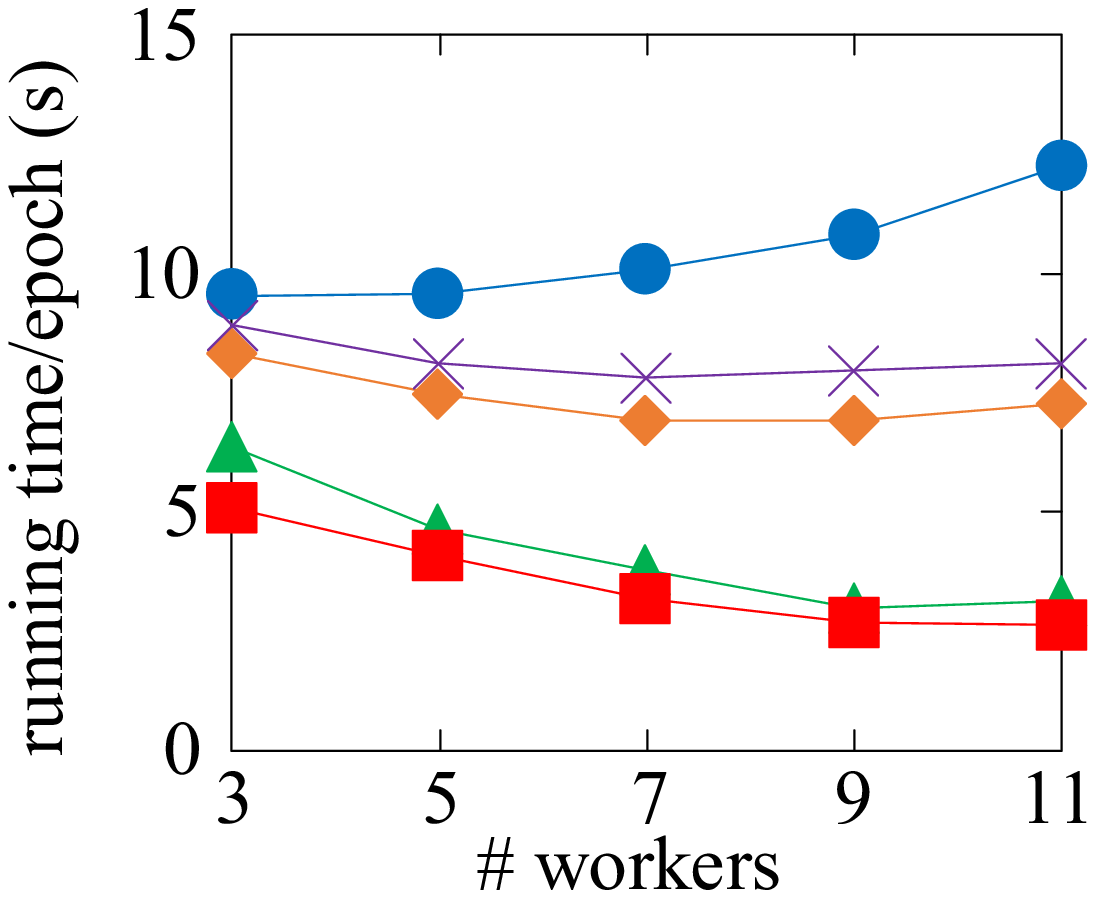}
  }
  \subfigure[\textit{KDD10}]{
   \label{fig:worker-KDD10}
   \includegraphics[width=0.235\textwidth]{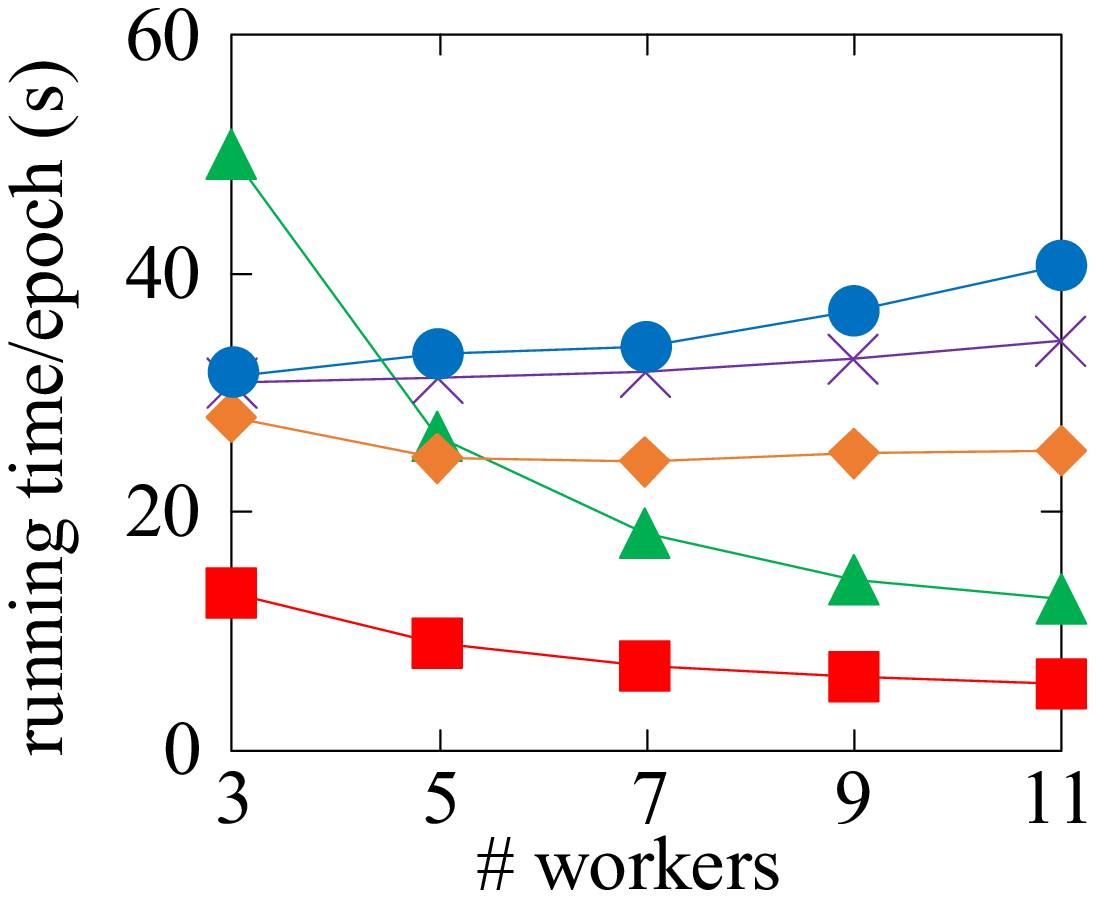}
  }
  \subfigure[\textit{KDD12}]{
   \label{fig:worker-KDD12}
   \includegraphics[width=0.235\textwidth]{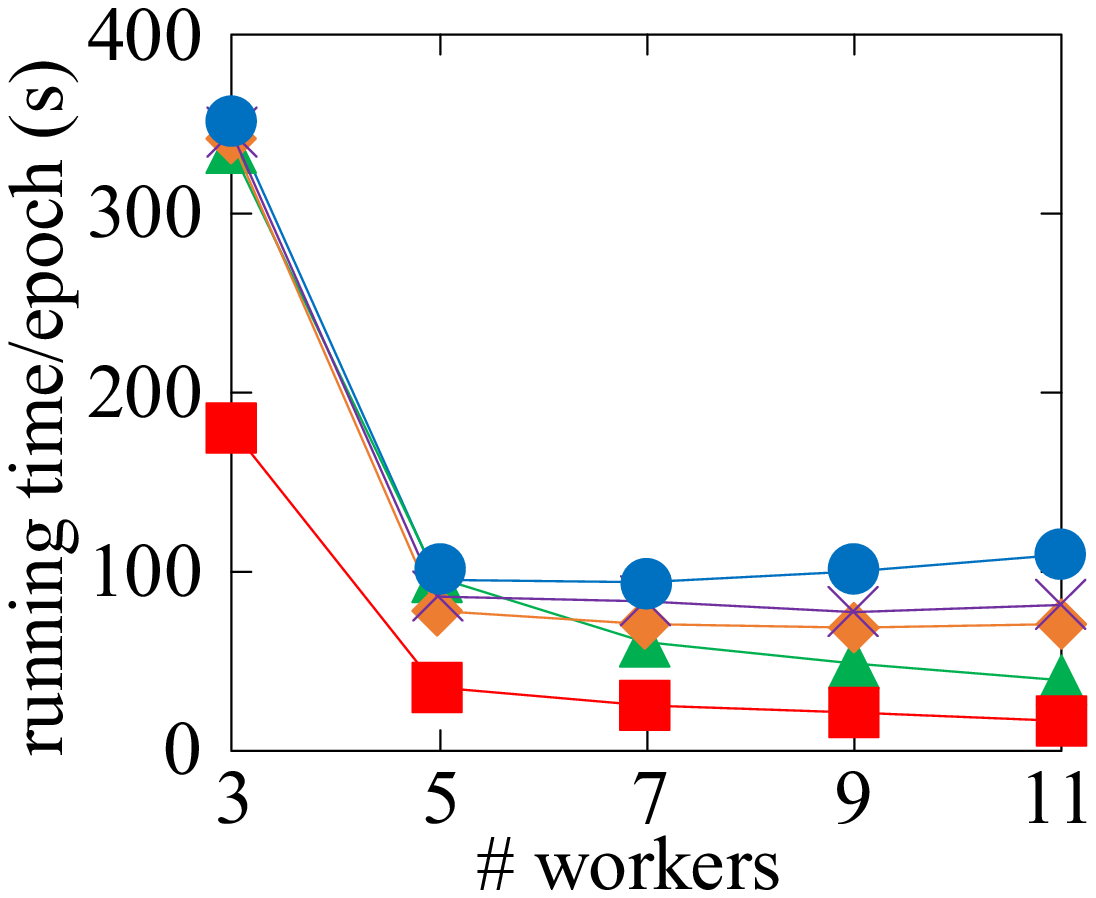}
  }
  \subfigure[\textit{WebSpam}]{
   \label{fig:worker-Web}
   \includegraphics[width=0.235\textwidth]{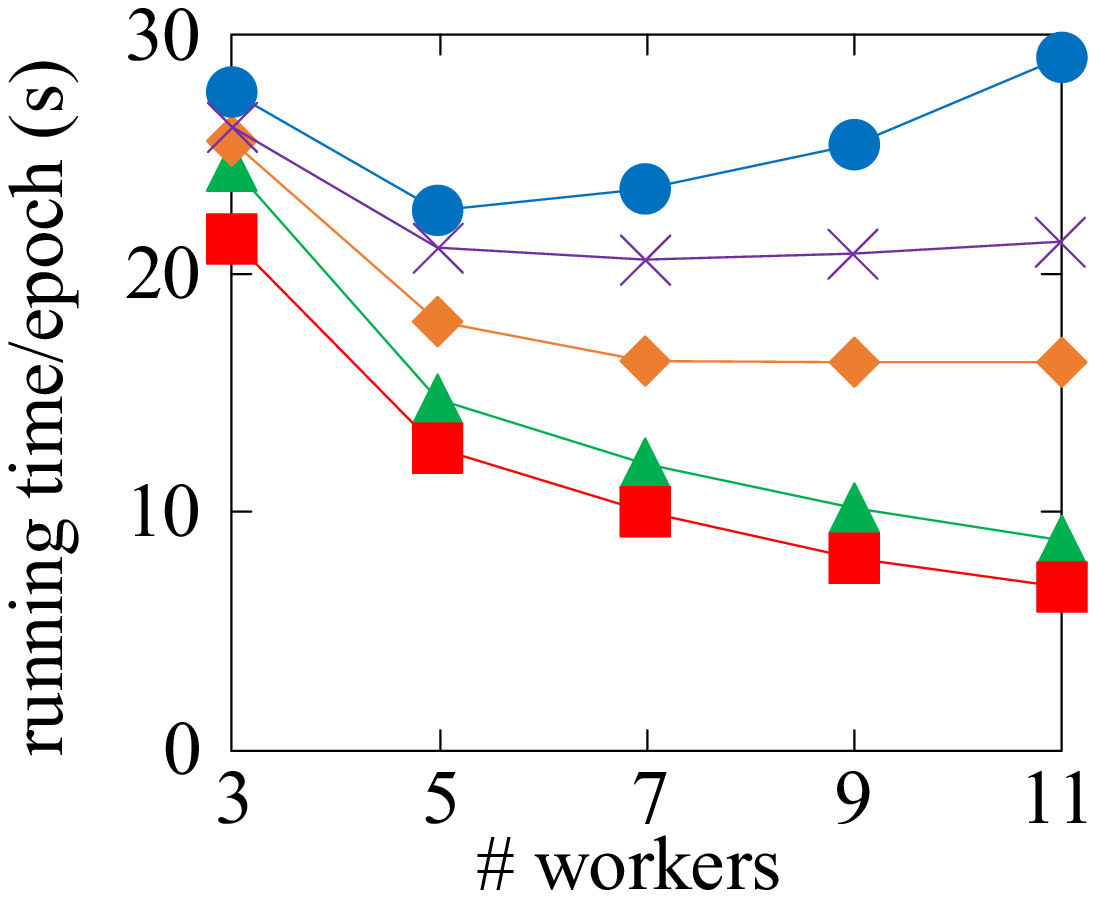}
  }
\vspace{-2mm}
\caption{Scalability comparison over SVM}
\label{fig:worker}
\end{figure*}

\begin{table*}[t]
% \center
\caption{The Performance of Convergence}
\vspace{-1mm}
\label{tab:conver}
\center
\begin{tabular}{|l|l|l|l|l|l|l|l|l|l|l|}
\hline
\multicolumn{11}{|c|}{\textbf{\textit{KDD12}}}                                                                                                                                                                                                              \\ \hline
\multirow{2}{*}{\textbf{Model}} & \multicolumn{2}{c|}{\textbf{FastSGD}}   & \multicolumn{2}{c|}{\textbf{Adam}}      & \multicolumn{2}{c|}{\textbf{LogQuant}}  & \multicolumn{2}{c|}{\textbf{SketchML}}  & \multicolumn{2}{c|}{\textbf{Top-$k$}}     \\ \cline{2-11}
                                 & \textbf{Min. loss} & \textbf{Cov. time} & \textbf{Min. loss} & \textbf{Cov. time} & \textbf{Min. loss} & \textbf{Cov. time} & \textbf{Min. loss} & \textbf{Cov. time} & \textbf{Min. loss} & \textbf{Cov. time} \\ \hline
Linear                               & 0.0214             & 348                & 0.0214             & 2969               & 0.0214             & 2765               & 0.0214             & 2007               & 0.0214             & 1189               \\ \hline
LR                              & 0.6924             & 437                & 0.6924             & 2638               & 0.6924             & 2392               & 0.6924             & 1836               & 0.6924             & 970                \\ \hline
SVM                           & 0.9968             & 331                & 0.9968             & 2441               & 0.9968             & 2550               & 0.9969             & 1871               & 0.9968             & 978                \\ \hline
\multicolumn{11}{|c|}{\textbf{\textit{WebSpam}}}                                                                                                                                                                                                            \\ \hline
\multirow{2}{*}{\textbf{Model}} & \multicolumn{2}{c|}{\textbf{FastSGD}}   & \multicolumn{2}{c|}{\textbf{Adam}}      & \multicolumn{2}{c|}{\textbf{LogQuant}}  & \multicolumn{2}{c|}{\textbf{SketchML}}  & \multicolumn{2}{c|}{\textbf{Top-$k$}}     \\ \cline{2-11}
                                 & \textbf{Min. loss} & \textbf{Cov. time} & \textbf{Min. loss} & \textbf{Cov. time} & \textbf{Min. loss} & \textbf{Cov. time} & \textbf{Min. loss} & \textbf{Cov. time} & \textbf{Min. loss} & \textbf{Cov. time} \\ \hline
Linear                               & 0.3174             & 110                & 0.3173             & 439                & 0.3176             & 365                & 0.3172             & 260                & 0.3173             & 133                \\ \hline
LR                              & 0.5884             & 102                & 0.5882             & 441                & 0.5892             & 314                & 0.5877             & 210                & 0.5885             & 122                \\ \hline
SVM                           & 0.6369             & 213                & 0.6371             & 838                & 0.6362             & 678                & 0.6435             & 479                & 0.6372             & 245                \\ \hline
\end{tabular}
\end{table*}

Fig.~\ref{fig:competitor} illustrates the convergence rate on four datasets.
Note that we only present the results of SVM to represent the performance on three ML models, since the results of the two others are similar.
%Due to the limitation of space, we only present the results of SVM to represent the performance on three ML models, while the results of the two others are similar.
It is observed that FastSGD achieves the fastest convergence rate, followed by Top-$k$, SketchML, and then LogQuant, while the basic Adam without the gradient compression is the worst. LogQuant and SketchML uses logarithmic quantization and sketch-based techniques respectively to compress the gradient. Thus, they converge faster than Adam. FastSGD and Top-$k$ perform better because they only transmit those gradient values with larger absolute values that can contribute more to the convergence of optimization algorithm based on the sparsification strategy. Furthermore, FastSGD squeezes the compression ratio by the presented encoders for both gradient values and keys.

Since FastSGD underestimates and discards the gradient values in the encoder, one may suspect whether FastSGD could converge and train the models successfully. Therefore, we conduct the experiments to demonstrate the convergence performance. We set the convergence condition of the optimization algorithm as if the change of validation loss is less than 1\%.

Table~\ref{tab:conver} lists the convergence performance of FastSGD, Adam, LogQuant, SketchML, and Top-$k$ on \textit{KDD12} and \textit{WebSpam} datasets. The metric is the minimal loss against converged time (seconds), separated by symbol `/'. We can observe that the four compressed algorithms could achieve almost the same model quality as Adam. Meanwhile, FastSGD converges faster than other algorithms. To be more specific, FastSGD converges faster than Adam by up to 8.5$\times$, faster than LogQuant by up to 7.9$\times$, faster than SketchML by up to 5.8$\times$, and faster than Top-$k$ by up to 3.4$\times$. As discussed in Section~\ref{sec:gv}, the logarithm quantization of FastSGD underestimates the gradient values. If a specific key index of a gradient is always underestimated, the convergence will be slow. Fortunately, the threshold filter would select different indices of the gradient to transmit in each training batch, while could alleviate this problem. Besides, the advantage of momentum and adaptive learning rate strategies in Adam could further reduce the sparsification error of threshold filter in FastSGD.

Finally, we compare the scalability of all the five algorithms. Fig.~\ref{fig:worker} shows the running time per epoch by varying the number of workers from 3 to 11 over SVM model on four datasets, while the results on the rest models and datasets are similar. As expected, the running time per epoch of all the competitors decreases as the number of workers ascends, except for Adam and LogQuant. For Adam, the running time increases with the number of workers on \textit{URL} and \textit{KDD10} datasets, while the running time first drops and then increases when the number of workers grows on \textit{KDD12} and \textit{WebSpam} datasets. For LogQuant, the running time increases with the number of workers on \textit{KDD10} dataset.

The reason is that, on one hand, the computation ability grows with the number of workers; on the other hand, the communication cost also increases with the number of workers.
%If the communication overhead becomes the bottleneck, the computation resource would experience idle times to wait for communication. It happens to Adam when the number of workers is larger than 3 on the small dataset \textit{URL} and the number of workers is larger than 5 workers for the big dataset \textit{WebSpam}.
The communication overhead becomes the bottleneck of the distributed system if the computation resource is sufficient.
%For Adam, it is sufficient to use 3 workers to train the small dataset \textit{URL} and use 5 workers for the big dataset \textit{WebSpam}.
The compression of gradients can alleviate the limitation of communication bottleneck by reducing the message size in the distributed system. Thus, FastSGD, Top-$k$, and SketchML could benefit from the number growth of workers. In addition, the higher the compression ratio (e.g., Top-$k$ or FastSGD), the better the scalability of the corresponding method.

\subsection{Performance over Neural Network}
In the aforementioned experiments, we have evaluated the performance of our proposed FastSGD with three generalized linear models. Furthermore, FastSGD can also be beneficial for distributed Neural Network models by compressing the transferring gradients. Next, we investigate the effectiveness of FastSGD over neural networks by training a Multilayer Perceptron (MLP) network on MNIST{\footnote{http://yann.lecun.com/exdb/mnist/}} dataset. The MLP network contains one input layer (size: 7 $\times$ 7), two fully connected layers (size: 100), and one output layer (size: 5). MNIST is a handwritten digits dataset, which consists of 60,000 training images and 10,000 testing images. The learning rate is set to 0.02.

\begin{figure}[t]
\centering
% \includegraphics[width=0.45\textwidth]{competitor-icon}\\
% \vspace{-1mm}
\includegraphics[width=0.49\textwidth]{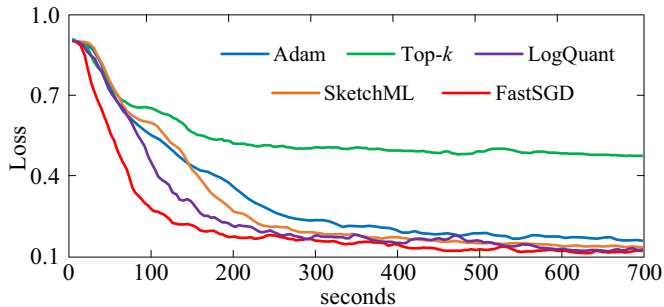}
\vspace{-2mm}
\caption{Convergence rate over neural network}
\label{fig:nn}
\end{figure}

Fig.~\ref{fig:nn} illustrates the convergence rate of FastSGD, Adam, LogQuant, SketchML, and Top-$k$. We can observe that Top-$k$ cannot perform well over the MLP network. This is because it discards too much gradient elements. Besides, FastSGD achieves both the fastest convergence rate and the smallest loss, followed by LogQuant, SketchML, and Adam. The reason is that FastSGD provides a lightweight gradient compression, which achieves a high compression ratio at a low cost while retain the main gradient information for optimization.

\section{Related Work}
\label{sec:related}

In this section, we overview three main categories of existing studies about the gradient compression for distributed ML, i.e., quantization, sparsification, and hybrid methods, respectively.

\subsection{Quantization Methods}
Quantization methods uses a small number of bits to represent each elements in the gradient.
Seide et al.~\cite{DBLP:conf/interspeech/SeideFDLY14} proposed the extreme form of quantization, 1-bit SGD, which quantizes the gradient element into a single bit by a user-defined threshold (0 by default). Dettmers~\cite{DBLP:journals/corr/Dettmers15} developed 8-bit Quantization to map each 32-bit gradient element into 8 bits. From those 8 bits, one bit is reserved for the sign of the number, three bits are used for exponent, while the rest four bits are used for mantissa.
%: 1 sign, 3 exponent, and 4 mantissa bits.
Alistarh et al.~\cite{DBLP:conf/nips/AlistarhG0TV17} presented QSGD, a family of compression schemes that allow a trade-off between the amount of communication and the running time. %by using random number generators.
TernGrad~\cite{DBLP:conf/nips/WenXYWWCL17} quantizes gradients to ternary levels (i.e., $\{-1,0.1\}$) with layer-wise ternarizing and gradient clipping to reduce the communication cost.
SignSGD~\cite{DBLP:conf/icml/BernsteinWAA18} transmits the sign of gradient elements by quantizing the negative ones to -1 and the others to 1.
Besides, Miyashita et al.~\cite{DBLP:journals/corr/MiyashitaLM16} proposed LogQuant, which uses the logarithmic data representation to eliminate bulky digital multipliers in convolutional neural networks. LogQuant can also be considered as a kind of quantization method to quantize gradients in log-domain.

Quantization methods can only achieve a limited compression ratio. Take the commonly used single-precision float (consisting of 32 bits) as an example. The maximal compression ratio of quantization methods is 32$\times$. Hence, we introduce a type of more aggressive compression technique, namely, sparsification, in the following.

\subsection{Sparsification Methods}
Researchers have confirmed that only a small number of gradient elements are needed to be aggregated during the gradient aggregate without hurting the model convergence and performance \cite{DBLP:conf/nips/WangniWLZ18, DBLP:conf/nips/StichCJ18, DBLP:conf/nips/JiangA18}. Following this idea, sparsification methods choose and transmit only a part of elements of the original gradient. The first sparsification method for large-scale machine learning is Truncated Gradient~\cite{DBLP:journals/jmlr/LangfordLZ09}, which induces sparsity in the weights of online learning algorithms.
Random-$k$~\cite{DBLP:journals/corr/KonecnyMYRSB16, DBLP:conf/nips/WangniWLZ18} selects $k$ elements in the gradient vector to communicate and update randomly.
Top-$k$~\cite{DBLP:conf/nips/AlistarhH0KKR18, DBLP:conf/ijcai/ShiZWTC19} finds the $k$ largest values in the gradient to transmit.
In contrast, Threshold-$v$~\cite{DBLP:conf/aaai/DuttaBA0SCK20} finds the elements whose absolute values are larger than a given threshold. Ivkin et al.~\cite{DBLP:conf/nips/IvkinRUBSA19} presented Sketched-SGD, which uses count-sketch~\cite{DBLP:journals/corr/MiyashitaLM16} to find large coordinates of a gradient vector to approximate the Top-$k$ gradient elements. DGC~\cite{DBLP:conf/iclr/LinHM0D18} picks up the top 0.1\% largest elements to compress the gradient while preserving the accuracy by momentum correction, local gradient clipping, momentum factor masking, and warm-up training.
Compared with DGC, SGC~\cite{DBLP:conf/dasfaa/SunSJ0LXW19} employs long-term gradient compensation to further improve the convergence performance.

\subsection{Hybrid Methods}
Hybrid methods combine both quantization and sparsification.
Storm~\cite{DBLP:conf/interspeech/Strom15} presented a hybrid method, which discards those gradient elements whose absolute values less than a pre-defined threshold and quantizes the others in a similar way as 1-bit SGD~\cite{DBLP:conf/interspeech/SeideFDLY14}.
Instead of a pre-defined threshold, Adaptive~\cite{DBLP:conf/sc/DrydenMJE16} determines the threshold dynamically by gradient samples.
Besu et al.~\cite{DBLP:conf/nips/0001DKD19} proposed Qsparse-local-SGD, which combines Top-$k$ or Random-$k$ sparsification with quantization.
3LC~\cite{DBLP:conf/mlsys/LimAK19} is another hybrid method that combines three techniques, i.e., 3-value quantization with sparsity multiplication, quartic encoding, and zero-run encoding.
Jiang et al.~\cite{DBLP:conf/sigmod/JiangFY018, DBLP:journals/vldb/JiangFYSC20} presented SketchML, which utilizes quantile sketch-based method to compress the sparse gradient consisting of key-value pairs.

Considering the gradients are sparse due to the original input data distribution and/or the sparsification technique, the key-value pair is a natural representation for the transferring gradient. However, most of the existing studies are only suitable for the dense gradient. Only SketchML~\cite{DBLP:conf/sigmod/JiangFY018, DBLP:journals/vldb/JiangFYSC20} that investigates the compression for both gradient keys and gradient values considers the sparse property for gradient. Nonetheless, it is limited to communication-intensive workload. This is because SketchML has a high computational cost of compression/decompression that could overwhelm the savings by the reduced gradient communication. In contrast, our proposed FastSGD aims at providing a lightweight and efficient compressed SGD framework for distributed ML.

\section{Conclusions}
\label{sec:conclusion}

In this paper, we propose FastSGD, a Fast compressed Stochastic Gradient Descent framework with lightweight compression/decompression cost for distributed machine learning.
%Consider the gradients are sparse originally and/or due to the sparsification compression strategy. FastSGD uses key-value pairs to represent the gradients.
%To achieve a high gradient compression ratio at a low cost,
FastSGD provides a compression framework for both the gradient keys and values in linear time complexity. For the gradient values, FastSGD offers a three-phase encoder, including the reciprocal mapper, the logarithm quantization, and the threshold filter, to compress the decimal gradient values into the small filtered gradient integers. For the gradient keys, an adaptive delta key binary encoder is provided to compress the gradient keys without producing any decoder error. Finally, we conduct extensive experiments on a range of practical ML models and datasets to demonstrate that FastSGD is more efficient than the state-of-the-art algorithms and scales well.

% \section*{Acknowledgments}

\balance
\bibliographystyle{abbrv}
\bibliography{FastSGD}
\balance

\balance

\end{document}